\documentclass[letterpaper]{article} 
\usepackage{aaai23}  
\usepackage{times}  
\usepackage{helvet}  
\usepackage{courier}  
\usepackage[hyphens]{url}  
\usepackage{graphicx} 
\usepackage{natbib}  
\usepackage{caption} 
\usepackage{algorithm}
\usepackage{algorithmic}
\usepackage{stfloats}
\usepackage{amsmath}
\usepackage{amssymb}
\usepackage{mathtools}
\usepackage{amsthm}
\usepackage{algorithm}
\usepackage{algorithmic}
\usepackage{amsfonts,comment,dsfont}
\usepackage[capitalize,noabbrev]{cleveref}
\usepackage{apxproof,array}
\newtheorem{theorem}{Theorem}

\newtheorem{lemma}{Lemma}

\newtheorem{remark}{Remark}
\newtheorem{example}{Example}
\usepackage[textsize=tiny]{todonotes}

\usepackage{xspace}

\newcommand{\bP}{\mathbb{P}}

\newcommand{\cB}{\mathcal{B}}

\newcommand{\cX}{\mathcal{X}}
\newcommand{\cA}{\mathcal{A}}
\newcommand{\cS}{\mathcal{S}}

\newcommand{\cU}{\mathcal{U}}
\newcommand{\cG}{\mathcal{G}}
\newcommand{\cD}{\mathcal{D}}

\newcommand{\cJ}{\mathcal{J}}

\newcommand{\cE}{\mathcal{E}}
\newcommand{\cN}{\mathcal{N}}
\newcommand{\cM}{\mathcal{M}}
\newcommand{\cH}{\mathcal{H}}

\newcommand{\cI}{\mathcal{I}}

\def\cA{{\mathcal{A}}} \def\cB{{\mathcal{B}}}  \def\cD{{\mathcal{D}}}
\def\cE{{\mathcal{E}}}  \def\cG{{\mathcal{G}}} \def\cH{{\mathcal{H}}}
\def\cI{{\mathcal{I}}} \def\cJ{{\mathcal{J}}} \def\cK{{\mathcal{K}}} 
\def\cM{{\mathcal{M}}} \def\cN{{\mathcal{N}}} \def\cO{{\mathcal{O}}} 
  \def\cS{{\mathcal{S}}} 
\def\cU{{\mathcal{U}}}  \def\cW{{\mathcal{W}}} \def\cX{{\mathcal{X}}}

\def\ba{{\mathbf{a}}}    
 \def\bg{{\mathbf{g}}}   
    \def\bo{{\mathbf{o}}}
 \def\bq{{\mathbf{q}}} \def\br{{\mathbf{r}}}

\def\bP{{\mathbf{P}}}

\newcommand{\bef}{\begin{figure}}
\newcommand{\eef}{\end{figure}}
\newcommand{\beq}{\begin{eqnarray}}
\newcommand{\eeq}{\end{eqnarray}}

\newcommand{\mab}{\texttt{MPMAB}\xspace}
\newcommand{\cmab}{\texttt{MPMAB-WA}\xspace}
\newcommand{\cmabucbma}{\texttt{MPMAB-WA-UCB-NR}\xspace}
\newcommand{\cmabucber}{\texttt{MPMAB-WA-UCB}\xspace}
\newcommand{\match}{\texttt{Learn2Match}\xspace}
\newcommand{\rank}{\texttt{Learn2Rank}\xspace}

\usepackage{newfloat}
\usepackage{listings}
\DeclareCaptionStyle{ruled}{labelfont=normalfont,labelsep=colon,strut=off} 
\floatstyle{ruled}
\newfloat{listing}{tb}{lst}{}
\floatname{listing}{Listing}
\title{Decentralized Stochastic Multi-Player Multi-Armed Walking Bandits}
\author{
 Guojun Xiong, Jian Li
}
\affiliations{
   SUNY-Binghamton University\\
    \{gxiong1,lij\}@binghamton.edu
}

\usepackage{bibentry}

\begin{document}

\maketitle

\begin{abstract}

Multi-player multi-armed bandit is an increasingly relevant decision-making problem, motivated by applications to cognitive radio systems.  Most research for this problem focuses exclusively on the settings that players have \textit{full access} to all arms and receive no reward when pulling the same arm.  Hence all players solve the same bandit problem with the goal of maximizing their cumulative reward. However, these settings neglect several important factors in many real-world applications, where players have \textit{limited access} to \textit{a dynamic local subset of arms} (i.e., an arm could sometimes be ``walking'' and not accessible to the player).  To this end, this paper proposes a \textit{multi-player multi-armed walking bandits} model, aiming to address aforementioned modeling issues. The goal now is to maximize the reward, however, players can only pull arms from the local subset and only collect a full reward if no other players pull the same arm.  We adopt Upper Confidence Bound (UCB) to deal with the exploration-exploitation tradeoff and employ distributed optimization techniques to properly handle collisions.  By carefully integrating these two techniques, we propose a decentralized algorithm with near-optimal guarantee on the regret, and can be easily implemented to obtain competitive empirical performance.

\end{abstract}

\section{Introduction}\label{intro}

The multi-armed bandit (MAB) framework has been widely adopted for studying sequential decision-making problems \cite{robbins1952some,lai1985asymptotically,auer2002finite,bubeck2012regret} in a variety of applications.  In a classic MAB setting, the decision maker chooses one arm from the set of $\mathcal{K}=\{1,\cdots,K\}$ arms at each time and receives a random reward according to unknown reward distributions.  The rewards of different arms are assumed to be independent and identically distributed over time.  The goal of the decision maker is to maximize the cumulative reward in the face of unknown mean rewards.

Recently, there has been an increased interest in studying the MAB in multi-player settings, dubbed as \mab, where the problem gets more intricate as $N$ independent decision makers (i.e., players) are involved.  At each discrete time $t$, each player selects one arm from $\mathcal{K}$, receives some feedback about this arm and possibly shares some ``information" with her neighbors.  Two popular settings have been widely studied: a collision setting, where a player collects the full reward from the selected arm only if no other players pull the same arm, as motivated by radio channel assignment in cognitive radios \cite{JEMP09}; and a collaborative setting, where players receive independent reward when they pull the same arm, and cooperatively solve a MAB, as motivated by sequential decisions in social networks \cite{landgren2016distributed}.  
In this work, we focus on the former setting, and simply refer to it as the \mab.

However, the basic model for \mab in most prior works assumes that players have \textit{full access} to all $K$ arms in each time.  This neglects several important factors of systems for many real-world applications, where each player can \textit{only} access a \textit{subset of arms} that \textit{dynamically} changes over time (i.e., an arm could sometimes be ``walking'' and not accessible to the player). 
For example, consider the problem of content placement in next-generation wireless networks (e.g., 5G/6G) \cite{andrews2014will} where $N$ cache-enabled base stations (players) serve a region where mobile users request for $K$ contents (arms), e.g., movies, videos, etc.  Users receive a large reward (e.g., a short latency) if the requested content is stored in the nearest base station, otherwise they are served by farther base stations with a small reward (e.g., a larger latency).   The base stations initially have no information about users' content requests and contents' global popularity since each base station only have access to a subset of contents due to its constrained cache size.  In reality, users' content requests are highly dynamic and hence each base station needs to repeatedly determine the subset of contents to be cached so as to maximize the total reward of serving users.   Another application is mobile edge computing \cite{ceselli2017mobile,farhadi2021service}, where edge clouds (arms) with computing resources form a shared resource pool, which can be allocated among user requests (players) that only have access to some edge clouds within the same geographical region. Additional real-world applications where players only have access to a subset of arms are presented in supplementary materials.

In this paper, we introduce a new bandit model in formalizing the \textit{walking arms} such that ``each player only accesses a subset of arms that dynamically changes over time".  Specifically, at time $t$, player $i$ only has access to a subset $\cS_i(t)\subseteq\mathcal{K}$ of arms, where $\cS_i(t)$ is changing over time.  As a result, we call the set of $\mathcal{K}$ arms as ``\textit{walking arms}'' and refer to the subset $\cS_i(t)$ as \textit{the local walking arm set}.  The goal is to find the optimal arm in $\cS_i(t)$ for each player $i\in\cN$ at each time $t$ to maximize the cumulative reward over a finite time horizon $T$.  However, player $i$ only observes a full reward if no other players pull the same arm.   We call this new bandit model as  \textit{``multi-player multi-armed walking bandits"} (\cmab).

To the best of our knowledge, this is the first work that integrates all three critical factors of multiple players, collisions and dynamic local walking arms into a unified \mab model.  However, the \cmab problem becomes much more challenging.  In particular, the dynamic local walking arms introduce an additional layer of complexity to the \mab problem that is already quite intricate.  This is because each player not only encounters a non-trivial tradeoff between \textit{exploration} (i.e., seeking better options) and \textit{exploitation} (i.e., staying with the currently-known best option) when attempting to maximizing the reward, but also is faced with \textit{a new dilemma} of how to manage the balance between maximizing the reward and avoiding collisions when players only receive feedback from a dynamic local subset of arms at each time.  

Though several known \mab algorithms can successfully handle the exploration-exploitation tradeoff, this new dilemma make existing arm elimination \cite{lykouris2018stochastic,gupta2021multi,boursier2019sic}, learning-to-rank \cite{combes2015learning,tibrewal2019multiplayer} and leader-follower  \cite{wang2020optimal,mehrabian2020practical} methods inapplicable in \cmab. 
In this paper, we make significant progress in this direction by extending the Upper Confidence Bound (UCB) \cite{auer2002finite} to deal with the exploration-exploitation tradeoff and employing distributed optimization techniques to properly handle collisions in the presence of walking arms.  This require careful integration of these techniques since the default optimal methods are incompatible with external randomness \cite{vernade2017stochastic,lykouris2018stochastic,madhushani2021one}.

Specifically, we study a ``networked information sharing'' setting, where all players are arranged in a network $\cG:=\{\mathcal{N}, \cE\}$, and each player has limited capacity for sharing information, e.g., their estimates of the arms' mean rewards with her neighbors in $\cG$, as inspired by the original idea of utilizing collisions to share sampled arm rewards in \mab settings \cite{boursier2019sic,shi2020decentralized}. To tackle the new dilemma in the presence of walking arms, we present a decentralized algorithm called \cmabucber, which is able to avoid collisions after sufficient exploration, in a decentralized manner, i.e., each player decides which arm to pull independently based on the local available information: the past observed rewards and collisions, along with the received information from neighbor players.  To achieve this, our high-level idea is to leverage shared information into exploitation to maximize reward from each player's perspective, which turns out to be a matching problem whose complexity grows exponentially with the number of players and arms.  To this end, we propose an efficient matching policy and a ranking policy, which assign different rankings to neighbor players so as to avoid collisions. We rigorously prove that a logarithmic growth of the regret is achievable for \cmabucber.   Note that our regret analysis is more challenging as traditional regret analysis becomes non-applicable here due to the integration of decentralized optimization methods for handling walking arms.

\section{Related Work}

As motivated by the cognitive radio channel assignment problem \cite{JEMP09}, the \mab problems have been extensively studied in different settings.  
There are two classes of algorithms for \mab.  The first class allows no information sharing among players, where players sense the presence of other players through experienced collisions \cite{anandkumar2011distributed}. 
The other class allows information sharing among players, e.g., directly sharing estimated mean rewards of arms \cite{liu2010distributed,kalathil2014decentralized,rosenski2016multi,bistritz2018distributed,besson2018multi,boursier2019sic,mehrabian2020practical,wang2020optimal,bubeck2020non,lugosi2021multiplayer,hanawal2021multi,pacchiano2021instance,shi2020decentralized}.  In particular, the regret guarantees for \mab were significantly improved in \cite{boursier2019sic} compared to the non-information sharing case.  However, the proposed SIC-MMAB needs to know the time horizon in advance and the exchange of reward estimations leading to the number collisions for communication grows large with $T$.  \cite{wang2020optimal,hanawal2021multi,shi2021heterogeneous} extended this model to a leader-follower framework with better regret guarantees.

However, all above literature assume that players have full access to all arms at each time while we consider a setting where players can only access a local subset of arms.  Furthermore, the local subset of arms is dynamically changing over time, which exhibits external randomness.  As a result, information sharing is necessary for \cmab to guarantee a near-optimal performance.  This is quite intuitive since there exists no universal ranking over arms across players due to the dynamic nature of \cmab.  This results in infinitely often collisions with an $\tilde{\mathcal{O}}(T)$ regret when all players independently pull arms in a greedy way.  We provide an intuitive example for further illustration along with additional related work discussions in supplementary materials.

\section{Problem Formulation}\label{sec:model}

We consider a stochastic multi-player multi-armed walking bandits (\cmab) with collisions setting with a set of $\cN=\{1,\cdots, N\}$ players, which are randomly distributed in a geographical region, and a set of $\cK=\{1,\cdots, K\}$ arms. Each arm $k$ is associated with a reward $X_k(t)$ at each discrete time $t=1,\cdots, T.$  The reward is a random variable on $(0,1]$ drawn independent and identically distributed (i.i.d.) from a certain distribution associated with arm $k$ with an unknown mean $\mu_k$.   Without loss of generalization (W.l.o.g.), we assume that $\mu_1>\mu_2>\cdots>\mu_K$.  In addition, in real-world applications, each player often has limited capability for information sharing, e.g., due to limited communication bandwidth. Thus we consider a networked setting where all players are arranged in a connected \textit{communication graph} $\cG:=\{\cN, \cE\}$ as the vertices.  Denote the neighbor players of player $i$ as $\cN_i=\{j|(i, j)\in\cE\}\cup\{i\}.$   

\textbf{Walking Arms with Collisions.} An arm could sometimes be ``walking'' and not accessible to a player.  Hence we call the set of $\cK$ arms as \textit{``walking arms''}.  Let $\mathcal{S}_i(t)\subseteq\cK$ denote the subset of available arms at time $t$ to player $i\in\cN$.  We refer to $\mathcal{S}_i(t)$ as \textit{``the local walking arm set''}, satisfying $\bigcup_i \mathcal{S}_i(t)=\mathcal{K}, \forall t.$ Since arms are walking, e.g., in a geographical area where players are located at (see our motivating examples in Introduction), we further assume that each arm can only be simultaneously accessed by neighbor players but not disjoint players in $\cG,$ i.e., $\cS_i(t)\cap \cS_j(t)=\phi$ if $j\notin \cN_i$. At time $t$, player $i$ can \textit{only} pull an arm from $\mathcal{S}_i(t)$, and \textit{only} observe a non-zero reward\footnote{There are other reward models for \mab settings, e.g.,  players can receive a degraded reward, or a full reward is only assigned to one  player when collisions occur \cite{liu2010distributed,liu2010decentralized2}. In this paper, we assume zero reward \cite{anandkumar2011distributed,besson2018multi} under collision for simplicity. However, our proposed model and algorithm can be easily generalized to other reward settings.} if no other neighbor players pull the same arm. Since our reward support is defined on $(0,1]$, i.e. $\mathbb{P}(X_k=0)=0$, the feedback scenarios referred to as \textit{collision sensing} and \textit{no sensing} settings  in \cite{boursier2019sic} are equivalent.

\textbf{Networked Information Sharing.}  
Inspired by the original idea of utilizing collisions to share sampled arm rewards \cite{boursier2019sic,shi2020decentralized}, each player $i\in\cN$ in our \cmab is able to share its local estimates of the arms' mean rewards with her neighbor players in $\cN_i$ at each time $t$.  Since players only have access to local walking arm sets in \cmab, and hence there exists no universal ranking over arms across players at each time.  Therefore, we further allow each player to share her local walking arm set with her neighbor players. 

\textbf{Policy.} A policy $\pi$ determines which arm each player will pull in each time.  We are interested in decentralized policies, where each player determines which arm to pull independently based on the available information to the player, including the past observed collisions, rewards, as well as possible information collected from neighbor players on local walking arm sets and reward estimates.  We denote the arm pulled by player $i$ at time $t$ as $a_i(t)$ under policy $\pi$, satisfying $a_i(t)\in\mathcal{S}_i(t)$.

\textbf{Regret.} We consider the performance measure of regret (in expectation) incurred by the set of $\cN$ players by pulling suboptimal arms under policy $\pi$ up to time $T$.  Since the local walking arm set $\mathcal{S}_i(t)\subseteq\cK, \forall i$ is varying over time $t$ and each player $i$ does not have full access to all arms in $\cK$, the optimal arms pulled by all players under the genie-aided algorithm that has knowledge of the true mean reward is not fixed.  This differs from existing works where the optimal expected reward can be simply achieved by pulling the best $N$ arms \cite{rosenski2016multi,besson2018multi,wang2020optimal}.  To this end, we denote the actions taken by all players under the genie-aided policy as $\ba^*(t):=[a^*_1(t), \ldots, a^*_N(t)],  \forall t$, satisfying 
\begin{align} 
\ba^*\!(t)\!=\! \arg\!\!\!\!\!\!\!\!\!\!\max_{\{\forall i:a_i(t)\in\cS_i(t)\}}\!\!\sum_{i=1}^{N}\!\mu_{ a_i(t)}\!\mathds{1}_{\{a_i(t)\neq a_j(t), \forall j\neq i, j\in\cN\}},
\end{align}
and the corresponding optimal expected reward as $R^*\triangleq\sum_{t=1}^{T}\sum_{i=1}^{N}\mu_{a^*_i(t)}(t)$. 
Then the regret up to time $T$ of policy $\pi$ is defined as
\begin{align}\label{eq:regret}
    \hspace{-0.2cm} R(T)\!\triangleq\! R^*\!-\!\mathbb{E}\!\left[\sum_{t=1}^{T}\!\sum_{i=1}^{N}X_{a_i(t)}(t)\mathds{1}_{\{a_i(t)\neq a_j(t), \forall j\neq i\}}\right].
\end{align}

\begin{remark}
The key difference of regret definition in \eqref{eq:regret} with that under full arm access setting (i.e., static arm setting) in prior works is the definition of $R^*$. Specifically, for a collision-free scenario \cite{martinez2019decentralized, madhushani2021one}, the $N$ players pull the best arm simultaneously and thus $R^*=TN\mu_1.$ For a collision setting \cite{anandkumar2011distributed, boursier2019sic,wang2020optimal}, the genie-aided algorithm assigns one of the $N$-best arms to each player and thus $R^*=T\sum_{i=1}^N\mu_i$.  The dynamic nature of our \cmab with local walking arm sets for each player brings external randomness and hence renders higher uncertainty for exploration and exploitation.  We will discuss its impact on the algorithm design and regret analysis in subsequent sections.

\end{remark}

\section{The \cmabucber Algorithm}

In this section, we consider \cmab under the above networked information sharing setting, and propose the \cmabucber algorithm to address the new dilemma faced by \cmab due to walking arms.

\subsection{Algorithm Overview}
Each player needs to resolve a tradeoff between exploration-exploitation and avoid collisions when attempting to maximize the reward: (i) pulling the arm with the largest estimated reward in her local walking arm set may contribute more to the total reward; and (ii) the neighbor players may share a similar estimation and local walking arm set, which may lead to a collision, and hence degrade the performance.  Exacerbating this dilemma is the fact that each player receives feedback from a dynamic local subset of arms at each time.  To resolve this dilemma, we leverage the shared information into the exploitation process to avoid collisions while maximizing the reward.
At each time $t$, \cmabucber starts with an information sharing process where player $i$ obtains the local walking arm sets $\{\cS_m(t), \forall m\in\cN\}$, and the local reward estimations $\tilde{r}_{i,k}(t), \forall k\in\cK$, from her neighbor players $\forall j\in\cN_i$.  Then \cmabucber alternates between exploration and exploitation as usual based on the past observed collisions and rewards.

\textbf{Information Sharing.}
At each time $t,$ player $i$ shares her local estimate of the mean reward $\tilde{r}_{i,k}(t)$, $\forall k\in\cK$ with her neighbor players $\forall j\in\cN_i$. Meanwhile, player $i$ receives the local estimates from her neighbors in $\cN_i$ and then updates her local reward estimates as follows: 
\begin{align}\label{eq:local-estimation-all}
\tilde{r}_{i,k}({t+1})=\sum\limits_{j\in \mathcal{N}_i} \!\tilde{r}_{j,k}(t)P_{i,j}\!+\! \hat{\mu}_{i,k}(t+1)\!-\!\hat{\mu}_{i,k}(t),
\end{align}
where $\bP=(P_{i,j})$ is a $N\times N$ non-negative matrix on the communication graph $\cG$ with $P_{i,j}\in[0,1]$, and $\hat{\mu}_{i,k}(t)$ is the empirical estimation of $\mu_k$ for player $i$ at time $t$, which will be specified later in~(\ref{eq:esti_reward}).  This update is analogous to the decentralized gradient method for decentralized optimization, where $\bP$ is referred to as the \textit{consensus matrix}\footnote{The easy-to-compute weights in (\ref{eq:consensus-matrix}) have been widely used in the decentralized optimization literature.  Our proposed model and algorithm are not restricted to (\ref{eq:consensus-matrix}) and can be easily generalized to other stochastic weights for $\bP$ \cite{Boyd05}.}, satisfying
\begin{align}\label{eq:consensus-matrix}
\begin{cases}
P_{i,i}= 1-\sum_{j\in\mathcal{N}_i} P_{i,j},\\
P_{i,j}=\frac{1}{\max\{|\mathcal{N}_i|, |\mathcal{N}_j|\}}, \quad\text{if $j\in\mathcal{N}_i$},\\
P_{i,j}=0,\quad \text{otherwise},
\end{cases}
\end{align}
with $\sum_{j=1}^N P_{i,j}=\sum_{i=1}^N P_{i,j}=1, \forall i, j.$
In other words, at each time $t$, player $i\in\cN$ computes a weighted average of the reward estimates of her neighbor players, and then corrects it by taking into account a stochastic approximation $\hat{\mu}_i(t+1)-\hat{\mu}_i(t)$ of her local reward estimate at time $t$.  As aforementioned, each player $i$ also shares her $\cS_i(t)$ so as to reach a consensus on the information of local walking arms set of the system, i.e., $\{\cS_m(t), \forall m\in\cN\}$ at each time $t$.

\textbf{Exploration.} 
The exploration of \cmabucber is based on the UCB exploration using all observations for each arm inside of the local walking arm set.  Essentially, each player runs UCB using the cumulative set of observations it has received.  We denote the number of times that player $i$ pulls arms $k$ by time $t$ as $I_{i,k}(t)$, in which collisions occur for $C_{i,k}(t)$ times.  Let $X_{i,k}(t)$ be the random reward received by player $i$ when pulling arm $k$ at time $t$.  Then the local reward estimation of $\mu_k, \forall k\in\cK$ for player $i$ at time $t$ is given as 
\begin{align}\label{eq:esti_reward}
    \hat{\mu}_{i,k}(t)=\frac{\sum\limits_{\tau=1}^t\mathds{1}_{\{a_i(\tau)=k, a_j(\tau)\neq k, \forall j\neq i\}}X_{i,k}(\tau)}{I_{i,k}(t)-C_{i,k}(t)},
\end{align}
where the numerator indicates the total rewards obtained by pulling arm $k$ without collisions, and the denominator denotes the corresponding times that no collisions occur.  
To accommodate the uncertainty of the local reward estimation $\tilde{r}_{i,k}(t)$ and follow the idea of UCB, we add a perturbed term to the estimated local reward in \eqref{eq:local-estimation-all} and define 
\begin{align}\label{eq:index_consensus}
   q_{i,k}(t)=\tilde{r}_{i,k}(t)+B_{i,k}(t),
\end{align}
with $B_{i,k}(t)$ being a function of $I_{i,k}(t)$ and $C_{i,k}(t)$.

\begin{remark}
Player $i$ often regards $q_{i,k}(t)$ in \eqref{eq:index_consensus} as an index of arm $k$, and pulls the arm with the largest index at time $t$ for exploitation in most prior works \cite{anandkumar2011distributed,boursier2019sic,wang2020optimal}.  However, this will inevitably cause a larger number of collisions since the local walking arm sets of neighbor players may share the same arm with the largest estimated reward. To alleviate collisions, learning-to-rank \cite{combes2015learning,tibrewal2019multiplayer} or leader-follower \cite{wang2020optimal,mehrabian2020practical} frameworks have been proposed where parsimonious exploration can be done by a single player (i.e., the leader) to find the best $N$ empirical arms, and then send this information to all other players (i.e., the followers).  However, these frameworks are based on the assumption that each player has full access to all arms, rendering them  inapplicable in \cmab, in which each player only has access to a dynamic local walking arm set.  As a result, there exists no such a best empirical arm set accessible for all players. To this end, a new exploitation strategy is needed to leverage the information received from neighbor players in the above information sharing process. 
\end{remark}

\begin{algorithm}[t]

	\caption{\cmabucber for player $i$ at time $t$} 
	\label{alg:procedure2}
	\textbf{Initialize:}
	The feasible arm sets for each player $\{\mathcal{S}_m(1), \forall m\in \mathcal{N}\}$; 
	the sample mean available at player $i$ $\{\hat{\mu}_{i,k}(1)=0, \forall k\in \mathcal{K}$, the local estimated reward $\{\tilde{r}_{i,k}(1)=0, \forall k\in \mathcal{K}$, and the statistics $\{q_{i,k}(1)=\infty,\forall k\in \mathcal{K}\}$;
	the number of pulls $\{I_{i,k}(1)=0, \forall  k\in \mathcal{K}\}$ and the number of collisions $\{C_{i,k}(1)=0, \forall k\in \mathcal{K}\}$.
	\begin{algorithmic}[1]
		\FOR{$t=1,...,T$}
       
        \STATE Share local walking arm sets among players to yield $\{\cS_m(t), \forall m\in\cN\}$;\\
        \STATE  Solve \eqref{eq:mathcing} by \match and select the arm indicated as $a_i^{*,i}$ by \rank; \\
        \STATE Update $I_{i,k}(t+1)$ and $C_{i,k}(t+1)$ according to~(\ref{eq:count_update});\\
         
         \STATE Update $\hat{\mu}_{i,k}(t+1)$ according to \eqref{eq:esti_reward}  and $\tilde{r}_{i,k}(t+1)$ according to \eqref{eq:local-estimation-all}; \\
         \STATE Update $q_{i,k}(t+1)$ according to \eqref{eq:index_consensus}.\\
		\ENDFOR
	\end{algorithmic}
\end{algorithm}

\textbf{Exploitation.} After sharing information with neighbor players and estimating the rewards of arms, each player $i$ determines which arm to pull at time $t$ from her local walking arm set $\cS_i(t)$. 
Since we are interested in decentralized decision makings, each player pulls one arm independently based on her local information.  As a result, each player $i$ has no information on the selected arms of her neighbor players.  To avoid collisions, player $i$ now leverages  $\{\cS_{m}(t), \forall m\in \mathcal{N}\}$ along with the estimated reward of $q_{i,k}(t)$ in \eqref{eq:index_consensus} to determine which arm to pull, instead of simply using $q_{i,k}(t)$ to pull the arm with the largest index value in $\cS_i(t)$.

Specifically, let $a_{i}^{m}(t)$ be the arm\footnote{Note that $a_{i}^{m}(t)$ is the arm pulled by player $m$ from the perspective of player $i$, which may not be the true arm pulled by player $m$ since players make decisions in a decentralized manner.} pulled by player $\forall m=1,\cdots, N$ from the perspective of player $i$ at time $t$. Denote $\ba_i(t)=\big[a_i^{1}(t), \ldots, a_i^{N}(t)\big]$ as the set of arms pulled by each player from the perspective of player $i$, and define the set containing all possible combinations as $\mathcal{U}_i(t)$ satisfying 
\begin{align}\label{eq:Uset}
    \mathcal{U}_i(t):=&\Big\{\ba_i(t)\Big|a_{i}^{m}(t)\in\mathcal{S}_{m}(t), \forall m=1,\cdots, N \Big\}.
\end{align}
Then player $i$ leverages the collected local walking arm sets $\{\cS_{m}(t), \forall m\in\mathcal{N}\}$, which are now embedded in $\mathcal{U}_i(t)$, together with her local estimated reward $\bq_i(t):=\{q_{i,k}(t), \forall k\in\cK\}$ to determine which arms all players should pull to maximize reward from her perspective.  This turns out to solving the following matching problem: 
\begin{align}\label{eq:mathcing}
 \max_{\ba_i(t)\in\mathcal{U}_i(t)}\sum_{m=1}^{N} q_{i, a_i^m(t)}(t)\mathds{1}_{\{a_i^m\neq a_i^n, \forall n\neq m, n=1,\cdots,N\}}.
\end{align}
Denote the optimal solution to~(\ref{eq:mathcing}) as $\ba_i^*(t)$.  Then player $i$ pulls arm $a_i^{*,i}(t)$ at time $t$.  Again, we note that $a_i^{*,m}(t)$ is the optimal arm that  player $\forall m=1,\cdots, N$ should pull at time $t$ by solving~(\ref{eq:mathcing}) from the perspective of player $i$.  Finally, player $i$ updates the indicators $I_{i,k}(t+1)$ and $C_{i,k}(t+1)$ based on the outcome of pulling arm $a_i^{*,i}(t)$ at time $t$, i.e., 
\begin{align}\label{eq:count_update}\nonumber
     I_{i,k}(t+1)&= I_{i,k}(t)+\mathds{1}_{\{a_{i}(t)=k\}},\\
     C_{i,k}(t+1)&=C_{i,k}(t)+\mathds{1}_{\{X_{i,k}(t)=0\}}.
\end{align}
We summarize our \cmabucber algorithm from  the perspective of any player $\forall i\in\mathcal{N}$ in Algorithm \ref{alg:procedure2}.

\subsection{\match}
To execute the exploration-exploitation process in Algorithm \ref{alg:procedure2}, player $i$ needs to solve the optimal matching problem in \eqref{eq:mathcing}, whose complexity grows exponentially with the number of  players $N$ and the number of arms in local walking arm set $\cS_{m}(t), \forall m=1,\cdots, N$, since $|\cU_i(t)|=\prod_{m=1}^{N}|\cS_{m}(t)|$.  To address this challenge, we now develop an efficient matching algorithm named \match to solve \eqref{eq:mathcing}, which is summarized in Algorithm~\ref{alg:Mathcing} from  the perspective of any player $\forall i\in\mathcal{N}$.  Since players receive no rewards when pulling the same arm, our approach to find an optimal $\ba_i^*(t)$ to maximize reward from the perspective of player $i$ over all other players  is straightforward: based on the local reward estimation $\bq_i(t)$ and all  players' local walking arm sets $\{\mathcal{S}_m(t), \forall m\in\mathcal{N}\}$, find $N$ ``feasible'' arms with the largest estimated reward that can be assigned to all players in $\cN$ to maximize \eqref{eq:mathcing}.

\begin{algorithm}[t] 
	\caption{\match for player $i$ at time $t$} 
	\label{alg:Mathcing}
	\textbf{Input:} $\cK,$ $\bq_i(t)$,  $\cU_i(t)$.\\
	\textbf{Ouput:}   $\cU^*_i(t).$ 
	\begin{algorithmic}[1]  
	  \STATE 
	  Let $\cA_i$ be a permutation on $\cK$ with a decreasing order based on the estimated reward $\bq_i(t)$, i.e.,  $q_{i,\cA_i^1}(t)\geq q_{i,\cA_i^2}(t)\geq\cdots\geq q_{i,\cA_i^{K}}(t)$;
	  \FOR{$h=1,2,...,N$}
		\STATE Add all players with local walking arm sets containing $\cA_i^1$ into $\cX_i^h$;
        
        \IF{$|\bigcup_{l=1}^h \cX_i^l|\geq h$} 
        \STATE $\cA_i=\cA_i\setminus\{\cA_i^1\}$,~$\cO_i(t)=\cO_i(t)\cup\{\cA_i^1\}$;
        \ELSE 
        \STATE $\cA_i=\cA_i\setminus\{\cA_i^1\}, \cX_i^h=\phi$, and $h=h-1$;
        \ENDIF

		\ENDFOR
		\STATE Let $\cS^*_{i,m}(t)=\cS_{m}(t)\cap \cO_i(t)$, $\forall m\in\cN$;
		\STATE Replace $\cS_{m}(t)$ by $\cS^*_{i,m}(t), \forall m\in\cN$  to obtain  $\cU^*_i(t)$.
	\end{algorithmic}
\end{algorithm}

Specifically, \match first constructs  a permutation on set $\cK$, denoted as $\cA_i$.  W.l.o.g., we order arms in $\cK$ in a decreasing order based on the estimated reward $\bq_i(t)$, and let $\cA_i^k$ denotes the $k$-th position\footnote{For abuse of notation, $\cA_i^k$ refers to the arm in $\cK$ with the $k$-th largest estimated reward from the perspective of player $i$.} in $\cA_i$ satisfying $q_{i,\cA_i^1}(t)\geq q_{i,\cA_i^2}(t)\geq\cdots\geq q_{i,\cA_i^{K}}(t).$ 
Based on this ordering, \match matches arms in $\cK$ to all players  by checking the arms with estimated rewards in a decreasing order defined by $\cA_i$, until finding $N$ feasible arms for all  players  at time $t$ (lines 2-9 in Algorithm~\ref{alg:Mathcing}).  For example, \match first checks the 1st position/arm $\cA_i^1$  with the largest estimate reward in $\cA_i$, and  adds all players whose local walking arm sets contain $\cA_i^1$ into $\cX_i^1$ (line 3 in Algorithm~\ref{alg:Mathcing}).  If the number of such players is no less than $h=1$, then arm $\cA_i^1$ is feasible and should be pulled by one player.  Thus \match adds it into the feasible arm set $\cO_i(t)$, and removes arm $\cA_i^1$ from $\cA_i(t)$, i.e., $\cA_i=\cA_i\setminus\{\cA_i^1\}$ (lines 4-5 in Algorithm~\ref{alg:Mathcing}).

Now suppose \match searches for the $h$-th arm to be added into  $\cO_i(t)$.  \match checks the arm in current $\cA_i^1$ and finds all players whose local walking arm sets contain $\cA_i^1$ and adds them into $\cX_i^h$.  If $|\bigcup_{l=1}^h \cX_i^l|\geq h$, i.e., the number of players that can pull the  $h$ arms in $\cO_i(t)\cup\{\cA_i^1\}$ is no less than $h$, and hence \match should remove the current $\cA_i^1$ arm from $\cA_i$ and put it into its feasible set $\cO_i(t)$ (lines 4-5  in Algorithm~\ref{alg:Mathcing}).  Otherwise, simple discard this arm since the number of arms in $\cO_i(t)$ is enough for all players in $|\bigcup_{l=1}^h \cX_i^l|$ to pull (line 7 in Algorithm~\ref{alg:Mathcing}).  As a result, \match ends up with a feasible arm set $\cO_i(t)$ that contains $N$ unique arms from $\cK$ that maximizes  \eqref{eq:mathcing}.  Finally, we update the local walking arm sets for all players and obtain the optimal arms pulled by all players at time $t$ as $\ba_i^*(t)$ and denote all possibilities as $\cU^*_i(t)$ (lines 10-11 in Algorithm~\ref{alg:Mathcing}).  The complexity for obtaining $\cO_i(t)$ and $\cU^*_i(t)$ is linear in the numbers of arms $K$ and players $N$. Since there may exist more than one optimal arm $a_i^{*,i}(t)$ that all maximize reward over all players from the perspective of player $i$, we next design a ranking policy named \rank  to assign different ranks to all  players to determine the unique arm pulled by player $i$ at time $t$.

\begin{algorithm}[t]
	\caption{\rank for player $i$ at time $t$} 
	\label{alg:rank}
	\textbf{Input:}  $\cU^*_i(t)$.

	\begin{algorithmic}[1]

		\STATE Construct $\cI_i:=\{a_i^{*,i}(t)\}$ and sort $\cI_i$ in a decreasing order based on $\bq_i(t)$;
		\STATE Construct $\cJ_i:=\{j,j\in\cN_i|\cI_i\subseteq\cS^*_{i,j}(t)\}$;
		\STATE Sort players $\forall j\in\cJ_i$ in a decreasing order according to their indices;
		\STATE Player $i$ pulls arm $\cI_i^{\beta_i}$ with $\beta_i$ being her ranking.
	\end{algorithmic}
\end{algorithm}

\subsection{\rank}
{Our key observation is that when there are $d$ different optimal arms $a_i^{*,i}(t)$, i.e., $|\cS_{i,i}^*(t)|=d$, then there must be $d$ players (including player $i$ herself) that are indifferentiable with these $d$ optimal arms.
Let $\cI_i:=\{a_i^{*,i}(t)\}$ be the set containing all  optimal arms that player $i$ can pull at time $t$. W.l.o.g, we order arms in $\cI_i$ in a decreasing order based on the estimated reward $\bq_i(t)$ such that $\cI_i^1\geq \cI_i^2\geq\ldots\geq\cI_i^{|\cI_i|}$ (line 1 in Algorithm \ref{alg:rank}). Then, \rank finds the set $\cJ_i:=\{j,j\in\cN|\cI_i\subseteq\cS^*_{i,j}(t)\}$ containing all neighbor players which can pull the optimal arms in $\cI_i$ as player $i$ (line 2 in Algorithm \ref{alg:rank}). In other words, players in $\cJ_i$ are indifferentiable with arms in $\cI_i$.  To avoid collisions, a simple rank strategy is to use players' indices.   Specifically, \rank sorts players in $\cJ_i$ in a decreasing order according to their indices, and then player $i$ pulls arm $\cI_i^{\beta_i}$ with $\beta_i$ being the ranking of player $i$ (lines 3-4 in Algorithm \ref{alg:rank}). This rank assignment associates each player in $\cJ_i$ with a unique ranking and hence can be used to avoid collisions. }

\begin{remark}
 We note that the idea of ranking players has also been adopted in recent works \cite{boursier2019sic,wang2020optimal}. However, all players are assumed to have full access to all arms at each time.  As a result, only one player needs to perform the ranking once and shares the universal ranking with all other players.  However, in our \cmab model, each player only has access to a local walking arm set that differs across players, and is dynamically changing over time. Hence there exists no universal ranking across players, making existing ranking methods \cite{boursier2019sic,wang2020optimal} inapplicable.  Finally, we provide an example in supplementary materials to illustrate the operations of our proposed \match and \rank policies. 
\end{remark}

\section{Performance Analysis}
In this section, we first analyze the performance of our \match and \rank policies, and then provide a finite-time analysis of \cmabucber.

\subsection{Collision Mitigation}
We first show that \match and \rank can be used to avoid collisions in \cmab. 

\begin{lemma}\label{lem:opt}
\match and \rank jointly provides an optimal solution to \eqref{eq:mathcing}, i.e., no collision occurs when the $\bq$-statistics are accurate.  
\end{lemma}

\begin{remark}
When local reward estimation $\bq$-statistics at each player are not accurate, players may pull sub-optimal arms and experience collisions, which incur regret (see Theorem~\ref{thm:A2} and Remark~\ref{remark:regret}).  When $\bq$-statistics are accurate (i.e., after a finite-time of exploration-exploitation), our \match and \rank jointly ensure an optimal solution to \eqref{eq:mathcing} without collisions.  Our proof consists of two steps.    First, based on the construction of $\cO_i(t)$ in \match using the expected estimated reward from the perspective of player $i$, and by contradiction, we show that $\cO_i(t)$ contains $N$ feasible arms, each pulled by one of the $N$ players which achieve the largest expected reward for \eqref{eq:mathcing}.  Second, since there may be more than one optimal arm to pull from the perspective of any player $i$, i.e., $|\cS_{i,i}^*(t)|>1$, and players determine which arm to pull in a distributed manner, collisions may occur if each player randomly pull an arm from $\cS_{i,i}^*(t)$.  To this end, \rank assigns a ranking to each player to determine the unique arm to pull from $\cS_{i,i}^*(t)$ and hence avoid collisions. 
\end{remark}


\subsection{Regret Analysis}\label{sec:regret}
We now provide a finite-time analysis of \cmabucber.  For ease of exposition, we define some additional notions.  
Let $V_{i,k}(t)$ be the number of times that arm $k\in\cK$ is only pulled by player $i\in\cN$ by time $t$, and denote $V_k(t):=\sum_{i=1}^N V_{i,k}(t)$.  Then the regret defined in \eqref{eq:regret} reduces to 
    $R(T)=R^*- \sum_{k=1}^K\sum_{i=1}^N\mu_k\mathbb{E}[V_{i,k}(T)].$
Furthermore, we define $I_k(t):=\sum_{i=1}^N I_{i,k}(t)$, where $I_{i,k}(t)$ is the number of times player $i$ pulling arm $k$ by time $t$ as defined earlier.  It is straightforward to see that $\sum_{k=1}^K I_k(T)=TN$.  We denote $\cK_b$ as the set containing arms with the largest $N$ mean reward, i.e.,  $\cK_b:=\{\mu_1, \mu_2, \ldots, \mu_N\},$ and let $\cK_{-b}=\mathcal{K}\setminus \cK_b$ contain the remaining arms.  Finally, let $C(T):=\sum_{k\in \cK_b}I_k(T)-V_{k}(T)$ be the number of collisions faced by players by pulling arms in $\cK_b$ by time $T$.

\begin{theorem}\label{thm:A2}
The regret of \cmabucber satisfies 
\begin{align*}
    &R(T)\leq \mu_1\Bigg(\max \left\{\sum_{k^\prime\in\cK_{-b}}\sum_{k\in\mathcal{K}\setminus\{k^\prime\}}  \frac{6\log T}{(\mu_k-\mu_{k^\prime})^2},NKL\right\}+\\
    &\max\left\{\sum_{k=1}^N\sum_{k^\prime=k+1}^K \!\frac{6\log T}{(\mu_k-\mu_{k^\prime})^2},NKL\right\}\!+\!\frac{\pi^2}{3}K(K+N)\!\Bigg),
\end{align*}
with $ B_{i,k}(t)=\sqrt{\frac{3\log t}{2NV_{i,k}(t)}}, \forall i\in\cN, k\in\cK$ and
$
  L=\min_t  3(1-\beta^N)^{t/24N(1+\beta^{-N})}\leq \frac{(1-\beta^N)}{48N(1+\beta^{-N})t},
$
where  $\beta$ is the smallest positive value of all consensus matrices, i.e., $\beta=\arg\min P_{i,j}$ with $P_{i,j}>0, \forall i,j\in\cN.$
\end{theorem}

\begin{remark}\label{remark:regret}
The first term corresponds to the regret incurred by pulling suboptimal arms during the exploitation. The second term is incurred by collisions on pulling the best $N$ arms when bad rankings caused by incorrect reward estimation, which dominates the regret due to low probability events of bad rankings from our \rank policy with good reward estimation.  The last term is the regret incurred by the exploration during the initial learning periods, which does not scale with the time horizon $T$ since after a finite time of exploration, all players learn the exact rank through our \rank policy and hence there would be no regret accumulating afterwards. 
\end{remark}

The regret of the first two terms scale with $\cO(K^2\log T)$ and $\cO(NK\log T)$, which is sub-logarithmic in time $T$ and matches the regret in existing works, e.g. \cite{anandkumar2011distributed,besson2018multi,boursier2019sic,wang2020optimal,mehrabian2020practical},
where all players are required to have full access to all arms in each time.  In contrast each player in our \cmab has the flexibility to access a dynamic subset of arms.    Though such flexibility of arm subsets regularly brings external randomness, it does not result in the multiplicative pre-factor that goes  with the time-dependent function in the regret to be higher than $KN$ in \cite{wang2020optimal, boursier2019sic}. For instance, the state-of-the-art algorithm SIC-MMAB \cite{boursier2019sic} achieves an asymptotically optimal regret of $\tilde{\mathcal{O}}(KN\log T)$ under the assumption that players have full arm access at each time.  In addition, it needs to know the time horizon in advance while our \cmabucber requires no knowledge on problem parameters.
The number of communication bits is upper bounded by $\mathcal{O}(N^2KT)$.  When the network is large, the communication may be predominant over the $\log T$, and hence it is interesting to further explore the joint effect of $K$ and $T$ instead of only considering asymptotic results in $T$, which largely remains exclusive in multi-player multi-armed bandit settings \cite{boursier2019sic}.

\begin{remark}
As discussed in Related Work, information sharing, in particular, the local walking arm sets, is necessary to guarantee a near-optimal performance for \cmab since players in \cmab can only access a local subset of arms, which is also dynamically changing over time.  In addition, we allow players to share their local estimates of the arms' mean rewards with their neighbor players in our \cmabucber algorithm as motivated by \cite{boursier2019sic,shi2020decentralized} which showed that such reward estimate sharing in the traditional \mab model improved regret guarantees compared to non-sharing case.  We now show that this is also true for \cmab model.  Specifically, we consider a variant of \cmabucber, where no reward estimate is shared among players, and call the corresponding policy as \cmabucbma.  We provide the detailed description of \cmabucbma and its regret analysis in supplementary materials.  As expected, \cmabucber attains an improved regret bound with a factor of $\mathcal{O}(N)$ compared to that of \cmabucbma.  This is intuitive  since player $i$ in \cmabucber also receives the reward estimation from her neighbors $\cN_i$ at each time, where $|\cN_i|<N,$ which can be regarded as a means to improve the exploration efficiency by a factor of $\mathcal{O}(N)$, i.e., an $\cO(N)$ decrease for the number of time steps needed to obtain the accurate statistics of arms.

\end{remark}

\section{Numerical Evaluations}
 \begin{figure}[t]
	\centering
	\begin{minipage}[b]{.225\textwidth}
    	\includegraphics[width=0.99\columnwidth]{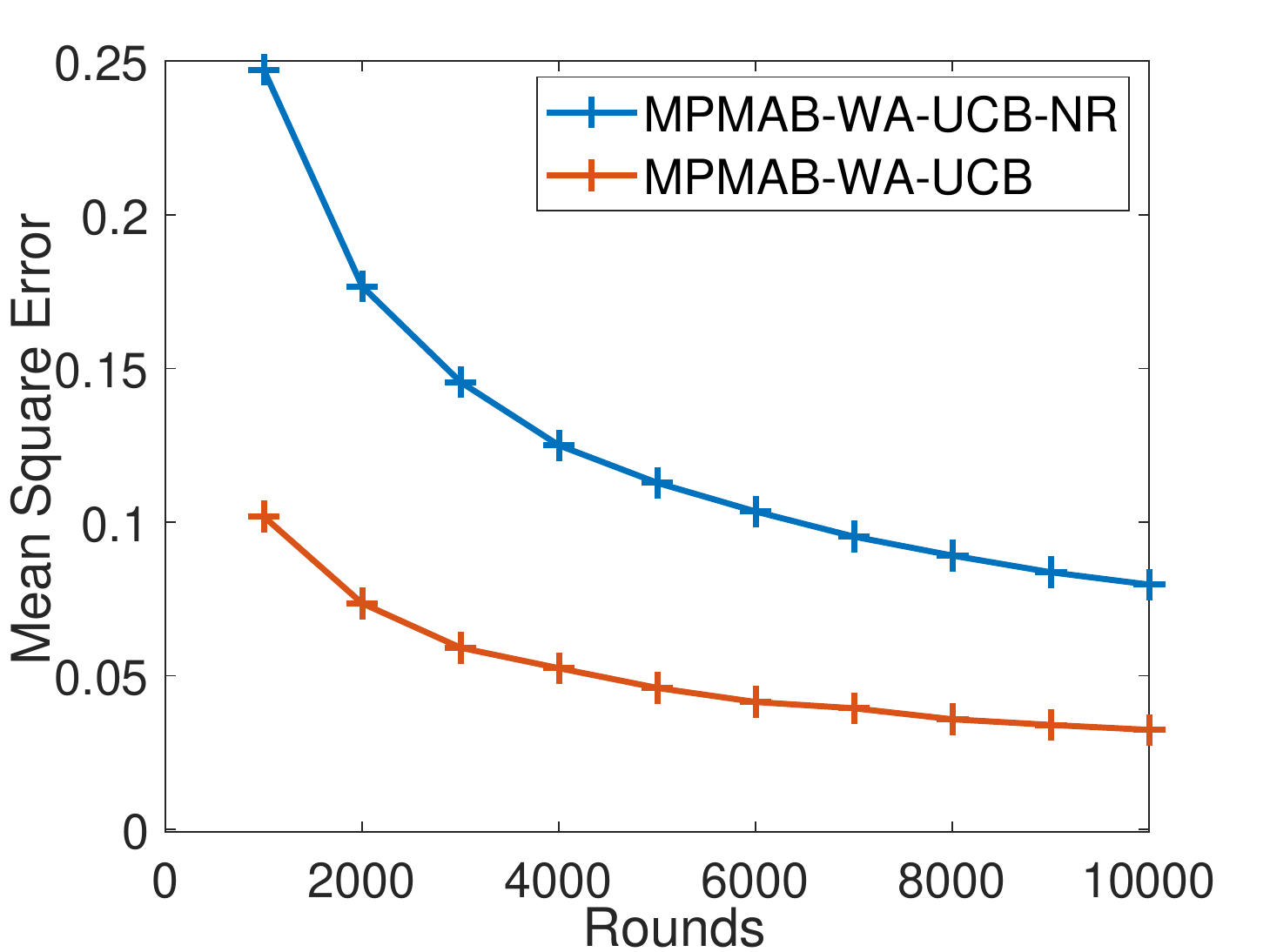}
		\caption{{MSE of mean reward.}}
	\label{fig:MSE}
	\end{minipage}
	\begin{minipage}[b]{.225\textwidth}
	    \includegraphics[width=0.99\columnwidth]{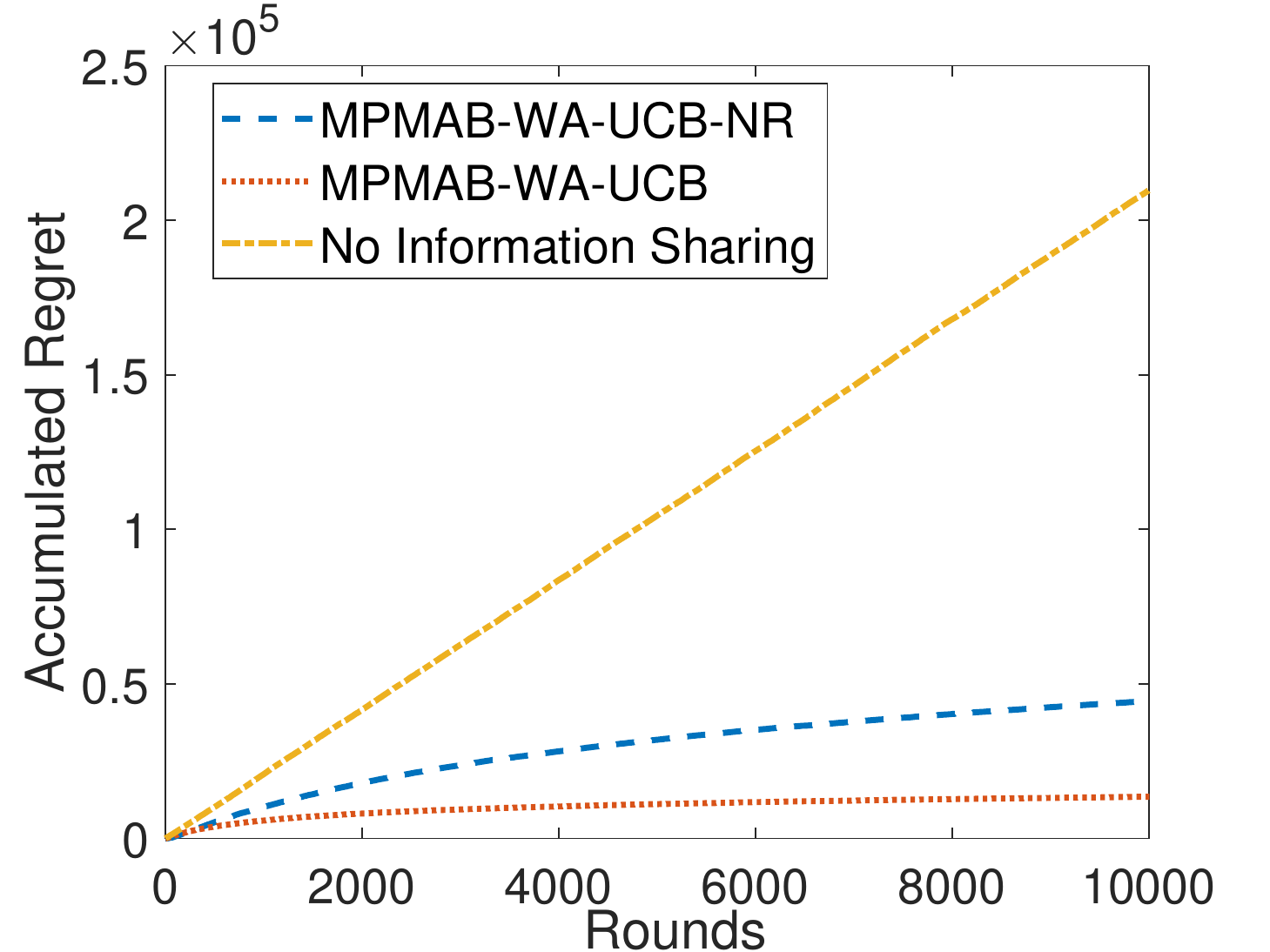}
		\caption{{Accumulated regret.}}
		\label{fig:Regret}
	\end{minipage}
\end{figure}

\textbf{Experiments on Constructed Instance.}  We consider $N=6$ players and $K=100$ arms with rewards drawn from Gaussian distributions with mean $\mu_k=0.06(101-k)$ and $\sigma_k=0.01(101-k)$.  Each player has three neighbor players in the communication graph $\cG$.  At each time, we randomly assign 25 arms to each player with neighbor players possibly sharing some arms.  All the regret and MSE values are averaged over 40 independent runs.  
Figure~\ref{fig:MSE} compares the mean-square-error (MSE) between each arm's true mean reward and estimated mean reward with our proposed algorithms over a time horizon of $T=10^4$ rounds.  It is clear that sharing estimated rewards with neighbor players as in \cmabucber  substantially improves the exploration efficiency compared to only sharing local walking arm sets as in \cmabucbma.  This advantage results in a lower regret as shown in Figure~\ref{fig:Regret}, which is consistent with our theoretical performance guarantees.  Finally, we observe that communication significantly improves the performance since communication is required to determine optimal matching and ranking to avoid collisions.  Its importance is especially pronounced when players only have access to a dynamic local walking arm set as considered in this paper.

\begin{figure}[t]
	\center
	\begin{minipage}[b]{.225\textwidth}
    	\includegraphics[width=0.99\columnwidth]{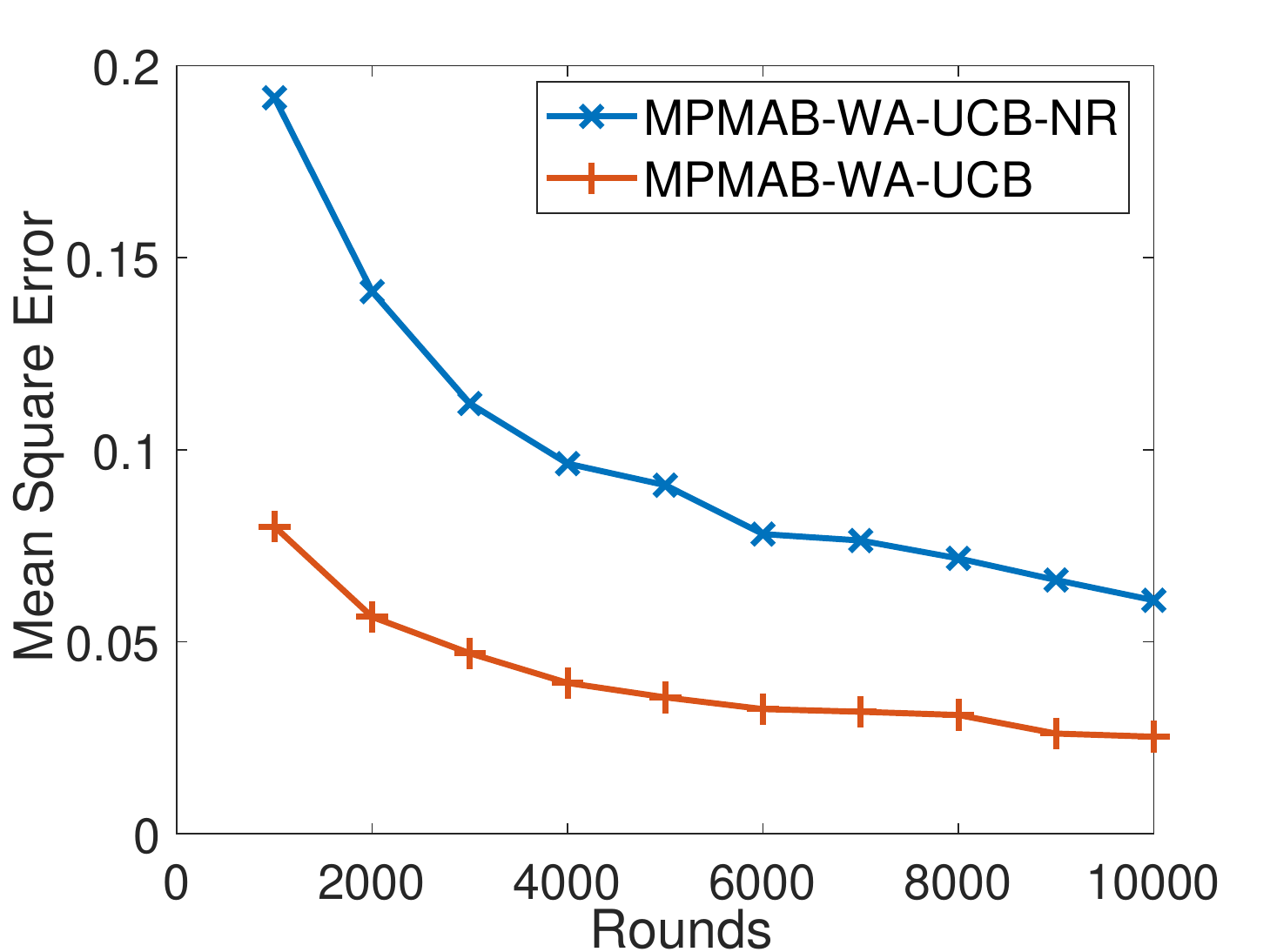}
		\caption{{MSE of mean reward in wireless downlink scheduling.}}
	\label{fig:MSE_DS}
	\end{minipage}
	\begin{minipage}[b]{.225\textwidth}
	    \includegraphics[width=0.99\columnwidth]{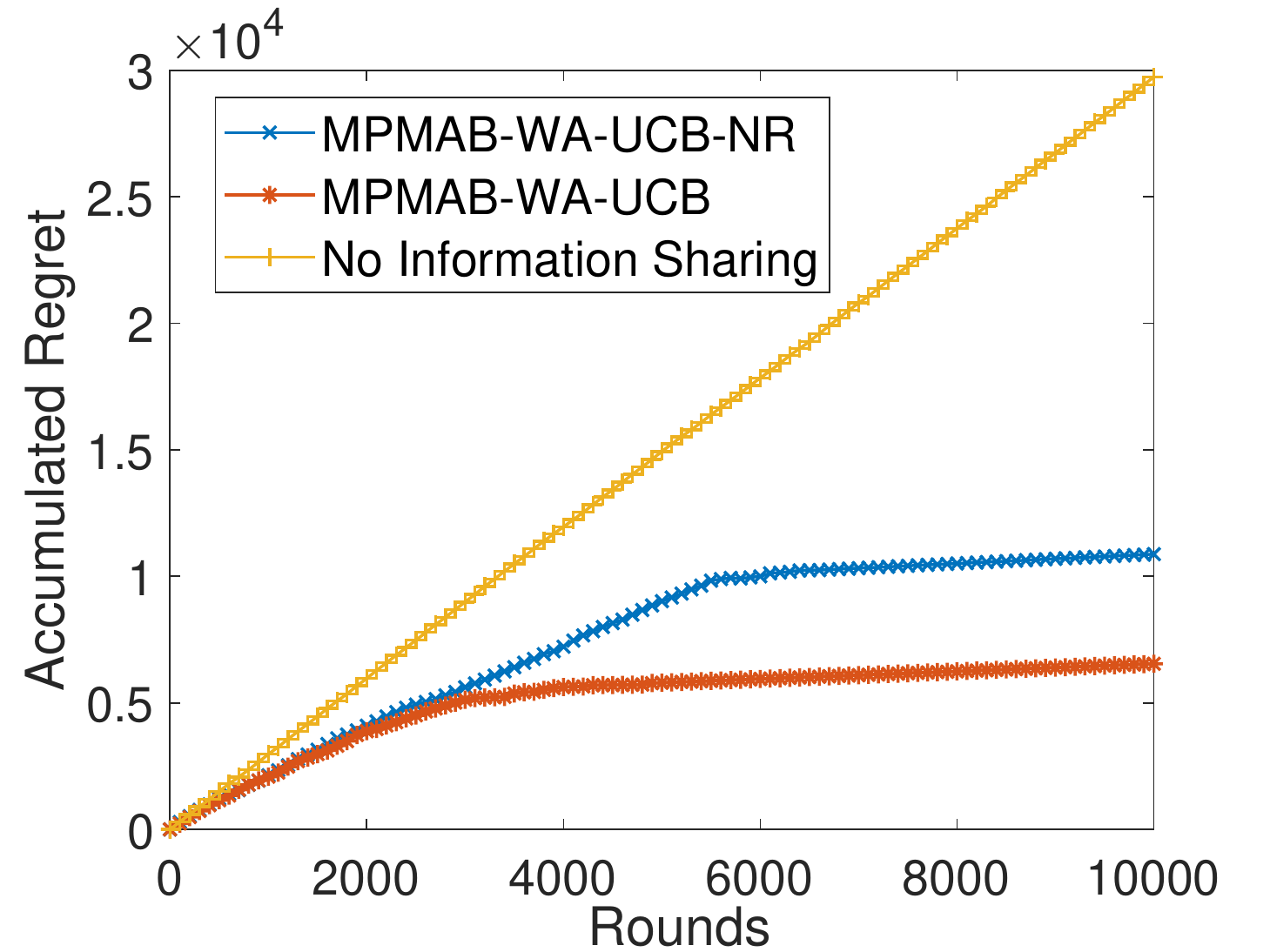}
		\caption{{Accumulated regret in wireless downlink scheduling.}}
		\label{fig:Regret_DS}
	\end{minipage}
\end{figure}

\noindent\textbf{Experiments on Wireless Downlink Scheduling.} We further consider a wireless downlink scheduling problem \cite{li2021efficient,li2019combinatorial} that fits into our \cmab model (see supplementary materials for details).  There are $N=6$ base stations (BSs) and $K=10$ walking users. Each BS covers a geographical region and each user randomly moves across the whole region with uniform distribution, i.e.,  each user moves into the region covered by BS $n$ with a probability ${1}/{N}$ at each time slot. BSs are connected via a ring, i.e., each BS has two neighbors. 
The rewards of serving users in each slot \cite{huang2021poster} are
i.i.d. drawn from Bernoulli distributions with mean rewards
$0.95, 0.9, 0.85, 0.8, 0.75, 0.7, 0.65, 0.6, 0.55, 0.5$.
All MSE and regret reported in Figures~\ref{fig:MSE_DS} and~\ref{fig:Regret_DS} are averaged over 40 independent runs, from which we draw the same conclusions as above.

\section{Conclusion}\label{sec:con}
In this paper, we studied the stochastic multi-player multi-armed bandits with collisions problem in the presence of walking arms, dubbed as \cmab.  This new framework integrates several critical factors of systems for many real-world applications.  In \cmab, each player only has access to a dynamic local walking arm set at each time, and only observes a full reward if no other players pull the same arm.  This introduced a new dilemma to manage the balance between maximizing the reward via exploration-exploitation, and avoiding collisions when players only receive feedback from a dynamic local walking arm set. To address this challenge, we considered a practical information sharing setting to coordinate players, and proposed a decentralized algorithm with theoretical guarantee on the regret.

\clearpage

\bibliographystyle{aaai23}
\bibliography{refs}

\begin{thebibliography}{58}
\providecommand{\natexlab}[1]{#1}

\bibitem[{Amani and Thrampoulidis(2021)}]{amani2021decentralized}
Amani, S.; and Thrampoulidis, C. 2021.
\newblock Decentralized Multi-Agent Linear Bandits with Safety Constraints.
\newblock In \emph{Proceedings of the AAAI Conference on Artificial
  Intelligence}, volume~35, 6627--6635.

\bibitem[{Anandkumar et~al.(2011)Anandkumar, Michael, Tang, and
  Swami}]{anandkumar2011distributed}
Anandkumar, A.; Michael, N.; Tang, A.~K.; and Swami, A. 2011.
\newblock Distributed algorithms for learning and cognitive medium access with
  logarithmic regret.
\newblock \emph{IEEE Journal on Selected Areas in Communications}, 29(4):
  731--745.

\bibitem[{Andrews et~al.(2014)Andrews, Buzzi, Choi, Hanly, Lozano, Soong, and
  Zhang}]{andrews2014will}
Andrews, J.~G.; Buzzi, S.; Choi, W.; Hanly, S.~V.; Lozano, A.; Soong, A.~C.;
  and Zhang, J.~C. 2014.
\newblock What will 5G be?
\newblock \emph{IEEE Journal on Selected Areas in Communications}, 32(6):
  1065--1082.

\bibitem[{Auer, Cesa-Bianchi, and Fischer(2002)}]{auer2002finite}
Auer, P.; Cesa-Bianchi, N.; and Fischer, P. 2002.
\newblock Finite-time analysis of the multiarmed bandit problem.
\newblock \emph{Machine Learning}, 47(2): 235--256.

\bibitem[{Besson and Kaufmann(2018)}]{besson2018multi}
Besson, L.; and Kaufmann, E. 2018.
\newblock Multi-player bandits revisited.
\newblock In \emph{Algorithmic Learning Theory}, 56--92. PMLR.

\bibitem[{Bistritz and Bambos(2020)}]{bistritz2020cooperative}
Bistritz, I.; and Bambos, N. 2020.
\newblock Cooperative multi-player bandit optimization.
\newblock \emph{Advances in Neural Information Processing Systems}, 33.

\bibitem[{Bistritz and Leshem(2018)}]{bistritz2018distributed}
Bistritz, I.; and Leshem, A. 2018.
\newblock Distributed multi-player bandits-a game of thrones approach.
\newblock \emph{Advances in Neural Information Processing Systems}, 31.

\bibitem[{Boursier and Perchet(2019)}]{boursier2019sic}
Boursier, E.; and Perchet, V. 2019.
\newblock SIC-MMAB: Synchronisation Involves Communication in Multiplayer
  Multi-Armed Bandits.
\newblock \emph{Advances in Neural Information Processing Systems}, 32:
  12071--12080.

\bibitem[{Bubeck and Cesa-Bianchi(2012)}]{bubeck2012regret}
Bubeck, S.; and Cesa-Bianchi, N. 2012.
\newblock Regret Analysis of Stochastic and Nonstochastic Multi-armed Bandit
  Problems.
\newblock \emph{Machine Learning}, 5(1): 1--122.

\bibitem[{Bubeck et~al.(2020)Bubeck, Li, Peres, and Sellke}]{bubeck2020non}
Bubeck, S.; Li, Y.; Peres, Y.; and Sellke, M. 2020.
\newblock Non-stochastic multi-player multi-armed bandits: Optimal rate with
  collision information, sublinear without.
\newblock In \emph{Conference on Learning Theory}, 961--987. PMLR.

\bibitem[{Ceselli, Premoli, and Secci(2017)}]{ceselli2017mobile}
Ceselli, A.; Premoli, M.; and Secci, S. 2017.
\newblock Mobile edge cloud network design optimization.
\newblock \emph{IEEE/ACM Transactions on Networking}, 25(3): 1818--1831.

\bibitem[{Chen et~al.(2021)Chen, Pasteris, Hajiesmaili, Lui, Towsley
  et~al.}]{chen2021cooperative}
Chen, Y.-Z.~J.; Pasteris, S.; Hajiesmaili, M.; Lui, J.; Towsley, D.; et~al.
  2021.
\newblock Cooperative Stochastic Bandits with Asynchronous Agents and
  Constrained Feedback.
\newblock \emph{Advances in Neural Information Processing Systems}, 34.

\bibitem[{Combes et~al.(2015)Combes, Magureanu, Proutiere, and
  Laroche}]{combes2015learning}
Combes, R.; Magureanu, S.; Proutiere, A.; and Laroche, C. 2015.
\newblock Learning to rank: Regret lower bounds and efficient algorithms.
\newblock In \emph{Proceedings of the 2015 ACM SIGMETRICS International
  Conference on Measurement and Modeling of Computer Systems}, 231--244.

\bibitem[{Dubey and Pentland(2020{\natexlab{a}})}]{dubey2020cooperative}
Dubey, A.; and Pentland, A. 2020{\natexlab{a}}.
\newblock Cooperative multi-agent bandits with heavy tails.
\newblock In \emph{International Conference on Machine Learning}, 2730--2739.
  PMLR.

\bibitem[{Dubey and Pentland(2020{\natexlab{b}})}]{dubey2020differentially}
Dubey, A.; and Pentland, A. 2020{\natexlab{b}}.
\newblock Differentially-private federated linear bandits.
\newblock \emph{Advances in Neural Information Processing Systems}, 33:
  6003--6014.

\bibitem[{Dubey and Pentland(2020{\natexlab{c}})}]{dubey2020kernel}
Dubey, A.; and Pentland, A. 2020{\natexlab{c}}.
\newblock Kernel methods for cooperative multi-agent contextual bandits.
\newblock In \emph{International Conference on Machine Learning}, 2740--2750.
  PMLR.

\bibitem[{Farhadi et~al.(2021)Farhadi, Mehmeti, He, La~Porta, Khamfroush, Wang,
  Chan, and Poularakis}]{farhadi2021service}
Farhadi, V.; Mehmeti, F.; He, T.; La~Porta, T.~F.; Khamfroush, H.; Wang, S.;
  Chan, K.~S.; and Poularakis, K. 2021.
\newblock Service placement and request scheduling for data-intensive
  applications in edge clouds.
\newblock \emph{IEEE/ACM Transactions on Networking}, 29(2): 779--792.

\bibitem[{Gupta et~al.(2021)Gupta, Chaudhari, Joshi, and
  Ya{\u{g}}an}]{gupta2021multi}
Gupta, S.; Chaudhari, S.; Joshi, G.; and Ya{\u{g}}an, O. 2021.
\newblock Multi-armed bandits with correlated arms.
\newblock \emph{IEEE Transactions on Information Theory}.

\bibitem[{Hanawal and Darak(2021)}]{hanawal2021multi}
Hanawal, M.~K.; and Darak, S. 2021.
\newblock Multi-player bandits: A trekking approach.
\newblock \emph{IEEE Transactions on Automatic Control}.

\bibitem[{Hillel et~al.(2013)Hillel, Karnin, Koren, Lempel, and
  Somekh}]{hillel2013distributed}
Hillel, E.; Karnin, Z.; Koren, T.; Lempel, R.; and Somekh, O. 2013.
\newblock Distributed exploration in Multi-Armed Bandits.
\newblock In \emph{Proceedings of the 26th International Conference on Neural
  Information Processing Systems-Volume 1}, 854--862.

\bibitem[{Hoeffding(1994)}]{hoeffding1994probability}
Hoeffding, W. 1994.
\newblock Probability inequalities for sums of bounded random variables.
\newblock In \emph{The collected works of Wassily Hoeffding}, 409--426.
  Springer.

\bibitem[{Huang, Hu, and Pan(2021)}]{huang2021poster}
Huang, Z.; Hu, B.; and Pan, J. 2021.
\newblock Poster: Multi-agent Combinatorial Bandits with Moving Arms.
\newblock In \emph{2021 IEEE 41st International Conference on Distributed
  Computing Systems (ICDCS)}, 1140--1141. IEEE.

\bibitem[{Jouini et~al.(2009)Jouini, Ernst, Moy, and Palicot}]{JEMP09}
Jouini, W.; Ernst, D.; Moy, C.; and Palicot, J. 2009.
\newblock Multi-armed bandit based policies for cognitive radio's decision
  making issues.
\newblock In \emph{2009 3rd International Conference on Signals, Circuits and
  Systems (SCS)}, 1--6. IEEE.

\bibitem[{Kalathil, Nayyar, and Jain(2014)}]{kalathil2014decentralized}
Kalathil, D.; Nayyar, N.; and Jain, R. 2014.
\newblock Decentralized learning for multiplayer multiarmed bandits.
\newblock \emph{IEEE Transactions on Information Theory}, 60(4): 2331--2345.

\bibitem[{Kanade, McMahan, and Bryan(2009)}]{kanade2009sleeping}
Kanade, V.; McMahan, H.~B.; and Bryan, B. 2009.
\newblock Sleeping experts and bandits with stochastic action availability and
  adversarial rewards.
\newblock In \emph{Artificial Intelligence and Statistics}, 272--279. PMLR.

\bibitem[{Kleinberg, Niculescu-Mizil, and Sharma(2010)}]{kleinberg2010regret}
Kleinberg, R.; Niculescu-Mizil, A.; and Sharma, Y. 2010.
\newblock Regret bounds for sleeping experts and bandits.
\newblock \emph{Machine Learning}, 80(2): 245--272.

\bibitem[{Kolla, Jagannathan, and Gopalan(2018)}]{kolla2018collaborative}
Kolla, R.~K.; Jagannathan, K.; and Gopalan, A. 2018.
\newblock Collaborative learning of stochastic bandits over a social network.
\newblock \emph{IEEE/ACM Transactions on Networking}, 26(4): 1782--1795.

\bibitem[{Lai and Robbins(1985)}]{lai1985asymptotically}
Lai, T.~L.; and Robbins, H. 1985.
\newblock Asymptotically efficient adaptive allocation rules.
\newblock \emph{Advances in Applied Mathematics}, 6(1): 4--22.

\bibitem[{Landgren, Srivastava, and Leonard(2016)}]{landgren2016distributed}
Landgren, P.; Srivastava, V.; and Leonard, N.~E. 2016.
\newblock Distributed cooperative decision-making in multiarmed bandits:
  Frequentist and bayesian algorithms.
\newblock In \emph{2016 IEEE 55th Conference on Decision and Control (CDC)},
  167--172. IEEE.

\bibitem[{Landgren, Srivastava, and Leonard(2018)}]{landgren2018social}
Landgren, P.; Srivastava, V.; and Leonard, N.~E. 2018.
\newblock Social imitation in cooperative multiarmed bandits: Partition-based
  algorithms with strictly local information.
\newblock In \emph{2018 IEEE Conference on Decision and Control (CDC)},
  5239--5244. IEEE.

\bibitem[{Lattimore and Szepesv{\'a}ri(2020)}]{lattimore2020bandit}
Lattimore, T.; and Szepesv{\'a}ri, C. 2020.
\newblock \emph{Bandit algorithms}.
\newblock Cambridge University Press.

\bibitem[{Li(2021)}]{li2021efficient}
Li, B. 2021.
\newblock Efficient learning-based scheduling for information freshness in
  wireless networks.
\newblock In \emph{IEEE INFOCOM 2021-IEEE Conference on Computer
  Communications}, 1--10. IEEE.

\bibitem[{Li, Liu, and Ji(2019)}]{li2019combinatorial}
Li, F.; Liu, J.; and Ji, B. 2019.
\newblock Combinatorial sleeping bandits with fairness constraints.
\newblock \emph{IEEE Transactions on Network Science and Engineering}, 7(3):
  1799--1813.

\bibitem[{Liu and Zhao(2010{\natexlab{a}})}]{liu2010decentralized2}
Liu, K.; and Zhao, Q. 2010{\natexlab{a}}.
\newblock Decentralized multi-armed bandit with multiple distributed players.
\newblock In \emph{2010 Information Theory and Applications Workshop (ITA)},
  1--10. IEEE.

\bibitem[{Liu and Zhao(2010{\natexlab{b}})}]{liu2010distributed}
Liu, K.; and Zhao, Q. 2010{\natexlab{b}}.
\newblock Distributed learning in multi-armed bandit with multiple players.
\newblock \emph{IEEE Transactions on Signal Processing}, 58(11): 5667--5681.

\bibitem[{Lugosi and Mehrabian(2021)}]{lugosi2021multiplayer}
Lugosi, G.; and Mehrabian, A. 2021.
\newblock Multiplayer bandits without observing collision information.
\newblock \emph{Mathematics of Operations Research}.

\bibitem[{Lykouris, Mirrokni, and Paes~Leme(2018)}]{lykouris2018stochastic}
Lykouris, T.; Mirrokni, V.; and Paes~Leme, R. 2018.
\newblock Stochastic bandits robust to adversarial corruptions.
\newblock In \emph{Proceedings of the 50th Annual ACM SIGACT Symposium on
  Theory of Computing}, 114--122.

\bibitem[{Madhushani et~al.(2021)Madhushani, Dubey, Leonard, and
  Pentland}]{madhushani2021one}
Madhushani, U.; Dubey, A.; Leonard, N.; and Pentland, A. 2021.
\newblock One more step towards reality: Cooperative bandits with imperfect
  communication.
\newblock \emph{Advances in Neural Information Processing Systems}, 34.

\bibitem[{Mart{\'\i}nez-Rubio, Kanade, and
  Rebeschini(2019)}]{martinez2019decentralized}
Mart{\'\i}nez-Rubio, D.; Kanade, V.; and Rebeschini, P. 2019.
\newblock Decentralized cooperative stochastic bandits.
\newblock In \emph{Proceedings of the 33rd International Conference on Neural
  Information Processing Systems}, 4529--4540.

\bibitem[{Mehrabian et~al.(2020)Mehrabian, Boursier, Kaufmann, and
  Perchet}]{mehrabian2020practical}
Mehrabian, A.; Boursier, E.; Kaufmann, E.; and Perchet, V. 2020.
\newblock A practical algorithm for multiplayer bandits when arm means vary
  among players.
\newblock In \emph{International Conference on Artificial Intelligence and
  Statistics}, 1211--1221. PMLR.

\bibitem[{Nedic and Ozdaglar(2009)}]{nedic2009distributed}
Nedic, A.; and Ozdaglar, A. 2009.
\newblock {Distributed Subgradient Methods for Multi-Agent Optimization}.
\newblock \emph{IEEE Transactions on Automatic Control}, 54(1): 48--61.

\bibitem[{Pacchiano, Bartlett, and Jordan(2021)}]{pacchiano2021instance}
Pacchiano, A.; Bartlett, P.; and Jordan, M.~I. 2021.
\newblock An Instance-Dependent Analysis for the Cooperative Multi-Player
  Multi-Armed Bandit.
\newblock \emph{arXiv preprint arXiv:2111.04873}.

\bibitem[{Robbins(1952)}]{robbins1952some}
Robbins, H. 1952.
\newblock Some aspects of the sequential design of experiments.
\newblock \emph{Bulletin of the American Mathematical Society}, 58(5):
  527--535.

\bibitem[{Rosenski, Shamir, and Szlak(2016)}]{rosenski2016multi}
Rosenski, J.; Shamir, O.; and Szlak, L. 2016.
\newblock Multi-player bandits--a musical chairs approach.
\newblock In \emph{International Conference on Machine Learning}, 155--163.
  PMLR.

\bibitem[{Sankararaman, Ganesh, and Shakkottai(2019)}]{sankararaman2019social}
Sankararaman, A.; Ganesh, A.; and Shakkottai, S. 2019.
\newblock Social learning in multi agent multi armed bandits.
\newblock \emph{Proceedings of the ACM on Measurement and Analysis of Computing
  Systems}, 3(3): 1--35.

\bibitem[{Shahrampour, Rakhlin, and Jadbabaie(2017)}]{shahrampour2017multi}
Shahrampour, S.; Rakhlin, A.; and Jadbabaie, A. 2017.
\newblock Multi-armed bandits in multi-agent networks.
\newblock In \emph{2017 IEEE International Conference on Acoustics, Speech and
  Signal Processing (ICASSP)}, 2786--2790. IEEE.

\bibitem[{Shi and Shen(2021)}]{shi2021federated}
Shi, C.; and Shen, C. 2021.
\newblock Federated multi-armed bandits.
\newblock In \emph{Proceedings of the 35th AAAI Conference on Artificial
  Intelligence (AAAI)}.

\bibitem[{Shi et~al.(2020)Shi, Xiong, Shen, and Yang}]{shi2020decentralized}
Shi, C.; Xiong, W.; Shen, C.; and Yang, J. 2020.
\newblock Decentralized multi-player multi-armed bandits with no collision
  information.
\newblock In \emph{International Conference on Artificial Intelligence and
  Statistics}, 1519--1528. PMLR.

\bibitem[{Shi et~al.(2021)Shi, Xiong, Shen, and Yang}]{shi2021heterogeneous}
Shi, C.; Xiong, W.; Shen, C.; and Yang, J. 2021.
\newblock Heterogeneous Multi-player Multi-armed Bandits: Closing the Gap and
  Generalization.
\newblock \emph{Advances in Neural Information Processing Systems}, 34.

\bibitem[{Szorenyi et~al.(2013)Szorenyi, Busa-Fekete, Hegedus, Orm{\'a}ndi,
  Jelasity, and K{\'e}gl}]{szorenyi2013gossip}
Szorenyi, B.; Busa-Fekete, R.; Hegedus, I.; Orm{\'a}ndi, R.; Jelasity, M.; and
  K{\'e}gl, B. 2013.
\newblock Gossip-based distributed stochastic bandit algorithms.
\newblock In \emph{International Conference on Machine Learning}, 19--27. PMLR.

\bibitem[{Tibrewal et~al.(2019)Tibrewal, Patchala, Hanawal, and
  Darak}]{tibrewal2019multiplayer}
Tibrewal, H.; Patchala, S.; Hanawal, M.~K.; and Darak, S.~J. 2019.
\newblock Multiplayer multi-armed bandits for optimal assignment in
  heterogeneous networks.
\newblock \emph{arXiv preprint arXiv:1901.03868}.

\bibitem[{Vernade, Capp{\'e}, and Perchet(2017)}]{vernade2017stochastic}
Vernade, C.; Capp{\'e}, O.; and Perchet, V. 2017.
\newblock Stochastic Bandit Models for Delayed Conversions.
\newblock In \emph{Conference on Uncertainty in Artificial Intelligence}.

\bibitem[{Vial, Shakkottai, and Srikant(2021)}]{vial2021robust}
Vial, D.; Shakkottai, S.; and Srikant, R. 2021.
\newblock Robust multi-agent multi-armed bandits.
\newblock In \emph{Proceedings of the Twenty-second International Symposium on
  Theory, Algorithmic Foundations, and Protocol Design for Mobile Networks and
  Mobile Computing}, 161--170.

\bibitem[{Wang et~al.(2020)Wang, Proutiere, Ariu, Jedra, and
  Russo}]{wang2020optimal}
Wang, P.-A.; Proutiere, A.; Ariu, K.; Jedra, Y.; and Russo, A. 2020.
\newblock Optimal algorithms for multiplayer multi-armed bandits.
\newblock In \emph{International Conference on Artificial Intelligence and
  Statistics}, 4120--4129. PMLR.

\bibitem[{Xiao, Boyd, and Lall(2006)}]{Boyd05}
Xiao, L.; Boyd, S.; and Lall, S. 2006.
\newblock {Distributed Average Consensus with Time-Varying Metropolis Weights}.
\newblock \emph{Automatica}.

\bibitem[{Xu, Tao, and Shen(2020)}]{xu2020collaborative}
Xu, X.; Tao, M.; and Shen, C. 2020.
\newblock Collaborative multi-agent multi-armed bandit learning for small-cell
  caching.
\newblock \emph{IEEE Transactions on Wireless Communications}, 19(4):
  2570--2585.

\bibitem[{Zhu and Liu(2021)}]{ZL21}
Zhu, J.; and Liu, J. 2021.
\newblock A Distributed Algorithm for Multi-Armed Bandit with Homogeneous
  Rewards over Directed Graphs.
\newblock In \emph{2021 American Control Conference (ACC)}, 3038--3043. IEEE.

\bibitem[{Zhu et~al.(2021)Zhu, Zhu, Liu, and Liu}]{zhu2021federated}
Zhu, Z.; Zhu, J.; Liu, J.; and Liu, Y. 2021.
\newblock Federated bandit: A gossiping approach.
\newblock In \emph{Abstract Proceedings of the 2021 ACM
  SIGMETRICS/International Conference on Measurement and Modeling of Computer
  Systems}, 3--4.

\end{thebibliography}

\appendix

\section{Additional Related Work}
We also note that cooperative stochastic bandits have also recently attracted a lot of attentions.  For stochastic bandits, decentralized cooperative estimation has been explored by running consensus protocol \cite{landgren2016distributed,landgren2018social,martinez2019decentralized} or via voting \cite{shahrampour2017multi}.  These algorithms often require agents to communicate real numbers to their neighbors.  A more realistic model has recently been developed in \cite{wang2020optimal}, which has been extended to several other settings \cite{dubey2020cooperative,dubey2020kernel,bistritz2020cooperative,amani2021decentralized,vial2021robust}.  In \cite{madhushani2021one}, the problem is further extended to imperfect communication.  Most of aforementioned work considered the settings in which agents have full access to all arms.  The dynamic local moving arm nature of our work is related to the category of sleeping bandits \cite{kanade2009sleeping,kleinberg2010regret,li2019combinatorial,amani2021decentralized} where some arms could be ``sleeping'' in some arounds.  Another line of work considered asynchronous bandits \cite{chen2021cooperative} in which agents only received feedback from a static local subset of arms but can still access all arms under a fully connected graph.  These basic models have also been extended to several other settings \cite{hillel2013distributed,sankararaman2019social,szorenyi2013gossip,kolla2018collaborative,shi2021federated,zhu2021federated,dubey2020differentially,chen2021cooperative}.  All aforementioned works
considered independently collected rewards across agents (i.e., no collisions), which stands in clear contrast to our work, since we consider \mab with collisions and moving arms.

\section{Motivating Application Examples}\label{sec:example-app}
\begin{figure}[tp]
    \centering
       \includegraphics[width=0.4\textwidth]{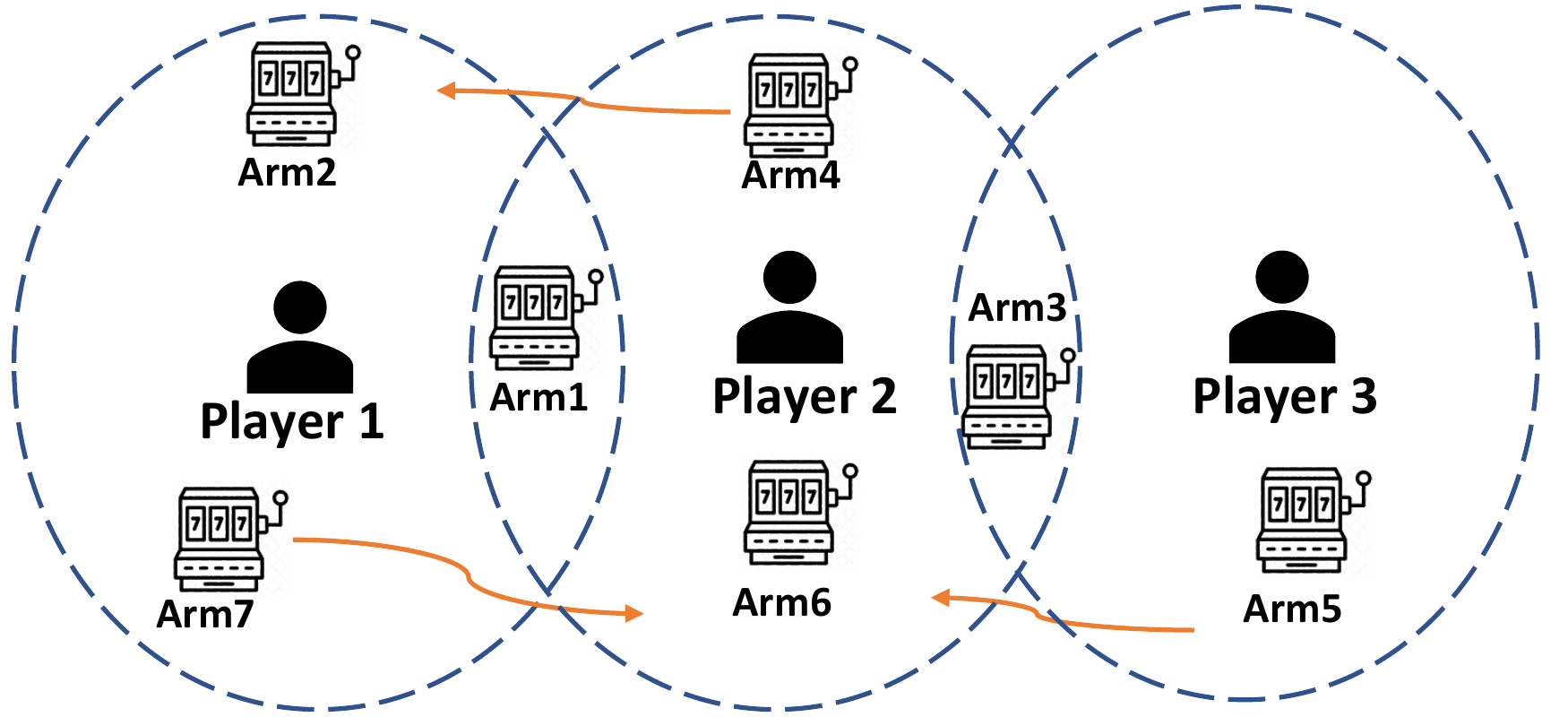}
\caption{An example of the \cmab system model with 3 players and 7 moving arms.} 
	\label{fig:1}
    \end{figure}

One key characteristic of our \cmab model is that each player only has access to a subset of arms at each time, which is called local moving arm set.  Furthermore, this local moving arm set is not fixed and dynamically changing over time.   In addition, we consider the scenario that each arm can only be simultaneously accessed by neighbor players but not disjoint players, as motivated by several real-world applications discussed below. 
An illustrative example of the system model is shown as in Figure \ref{fig:1}. There are 3 players and 7 moving arms in the system. At the current time, the local moving arm set of player 1 contains arms 1, 2, 7, of player 2 contains arms 1, 3, 4, 6, and of player 3 contains 3, 5, respectively. The collision will only occur between neighbor players, for instance, between player 1 and player 2, or player 2 and player 3. Since player 1 and player 3 are not neighbors to each other, they have disjoint available moving arm sets, and thus there is no collision between them. This is an important factor of systems in real-world applications but neglected in the classic \mab with collisions model where players are assumed to have full access to all arms at each time.  We provide two real-world scenarios that can be modeled with our proposed \cmab with some simplifications.  The first example is a wireless downlink scheduling problem and the second example is the small-cell caching scheduling problem.  
For ease of argument, we assume each BS only serves one client or caches one file at each time (i.e., capacity limited to one).  

$\bullet$ \textbf{Wireless Downlink Scheduling \cite{li2021efficient}.} Consider a real-time traffic scheduling problem in a wireless network with $N$ BSs and $K$ clients. Each individual BS only covers a sub-region and all BSs jointly cover the whole region. The sub-region covered by one player may overlap with the sub-regions of nearby players. We call those players which have overlapped area coverage neighbor players.   
The $K$ clients are moving
across the whole area and each client can only be connected to those BSs which cover the area the client currently lies in. Thus, the available clients connected to each BS are different over time.  Assume that time is slotted and  a scheduling cycle consists of $T$ consecutive time slots.
At the beginning of each scheduling cycle, every client generates a request signal to all available BSs. 
Each BS selects one client to serve.  Any request can only be served by one BS  to avoid collisions.  If a user request is successfully served, a utility (a measure of the value for the service) is generated. The utility
of each request from the same client can be assumed to be
random variables and the mean utility value for each client
is unknown in advance. The objective of the scheduling is to
gain more utility in expectation as much as possible for the whole network. The wireless downlink scheduling problem can be formulated as a standard multi-player bandits in the presence of
moving arms and collisions.  The $N$ BSs correspond to the $N$ players and $K$ moving clients represents the $K$ moving arms. At each time, each BS $n, \forall n\in\{1,2,\ldots,N\}$ selects one client $k\in \cS_n(t)$ to maximize the total accumulated reward in a finite-horizon $T$.

$\bullet$ \textbf{Small-Cell Caching Scheduling \cite{xu2020collaborative}.} 
Consider a cellular network composed of $N$ small-cell base
stations (BSs),  and $K$ different users moving across the area covered by BSs with each user requiring a specific file to download. Each BS can cache a file for one user at each time. 
Assume time is slotted and in each  slot, a user can only be connected to one BS and download the file from that BS. If two BSs simultaneously cache the same file for one user, the reward will be only assigned to the BS which is geologically closer to the user and the other client receives zero reward. The user will move from one region to another over time. The considered small-cell caching problem can be formulated as multi-player bandits problem in  the presence of moving arms and collision by considering the BSs as players and users as $arms$.  At each time, each BS selects one file to cache and the goal is to 
maximize the accumulated reward in a finite-horizon $T$.

\section{Motivating Example for Communication}\label{sec:related-app}
Communication is necessary to guarantee a near-optimal performance for the dynamic moving arm setting. In particular, the communication information will be used to determine optimal matching to avoid collisions.  This is quite intuitive since otherwise all players will independently pull arms in a greedy way, i.e., simply pull the best arm in her local arm set.  This results in infinitely often collision with a $\mathcal{O}(T)$ regret, even if the local moving arm set is fixed.  An intuitive example is presented below, and we also numerically verify in Figure~2 and Figure~4 in the numerical evaluation section of the main paper.  
\begin{example}
Suppose there are $3$ players and $5$ arms. At current time $t$, player $1$ has the local moving arm set $\cS_1(t)=\{1, 3\},$  player $2$ has the local moving arm set $\cS_2(t)=\{1, 2, 4\}$, and player $3$ has the local moving arm set $\cS_3(t)=\{2, 5\}$. Specifically, $\mu_1>\mu_2>\mu_3>\mu_4>\mu_5$.  Assume that each player has sufficiently explored all arms with perfect knowledge about all arms.  However, each player has no information about the local moving arm sets of other players in the system.  Under the above setting, the optimal pulling strategy at time $t$ is $a_1^*(t)=3, a_2^*(t)=1,$ and $a_3^*(t)=2$, i.e., player 1 pulls arm 3, player 2 pulls arm 1 and player 3 pulls arm 2. However, each player has no information about this optimal policy since we are interested in decentralized algorithm and each player makes decisions independently based on the local available information.  To this end, each player randomly selects one arm.  There are total $12$ different pulling strategies for the three players.  Hence, it causes regret with probability $\frac{11}{12}$.  At time $t+1$, $\cS_1(t+1)=\{1,2\},\cS_2(t+1)=\{2,3,5\},$ and $\cS_3(t+1)=\{2,4\}$. The corresponding optimal policy is  $a_1^*(t)=1, a_2^*(t)=3,$ and $a_3^*(t)=2$. Since player 2 has no information for $\cS_3(t+1)$ and player $3$ has no information of $\cS_2(t+1)$, they randomly select arms from their own local moving arm sets, respectively and thus cause regret with probability $\frac{5}{6}$.  As a result,  the accumulated regret is $\mathcal{O}(T)$ after a finite-horizon $T$ if no information on the local moving arm sets is shared among players since each player only has access to a local moving arm set, which is dynamically changing over time, in contrast to the assumption that players have access to all arms in most prior works.  
\end{example}


\section{More discussion about \match and \rank}

The major complexity of these procedures comes from solving the matching problem in \eqref{eq:mathcing}. Because the cardinality of the set $\cU_i(t)$ is the product of the cardinality of each set $\cS_{\cH_i(m)}(t), \forall m\in [|\cH_i|]$, i.e., $|\cU_i(t)|=\prod_{m=1}^{|\cH_i|}|\cS_{\cH_i(m)}(t)|$, which exponentially increases as the number of neighbor players $|\cH_i|$ and the number of arms for each set $\cS_{\cH_i(m)}(t), \forall m\in [|\cH_i|]$. To reduce the complexity, we design an efficient matching algorithm in Algorithm \ref{alg:Mathcing} for \eqref{eq:mathcing}. It turns out that the complexity for searching $\bo_i$ is linearly with $|\cA_i|$ and the complexity for recovering $\cU_i(t)$ from $\bo_i$ is less than $factorial(|\cH_i|)$.

\subsection{Optimality of \match}


\subsection{An Illustrative Example for \match and \rank}\label{sec:ilustration-app}
We provide an illustrative example to explain the operations of our proposed \match (Algorithm~\ref{alg:Mathcing}) and \rank (Algorithm~\ref{alg:rank}) algorithms, and the necessity of designing these two algorithms to avoid collisions as many as possible. 
\begin{figure}[t]
    \centering
       \includegraphics[width=0.45\textwidth]{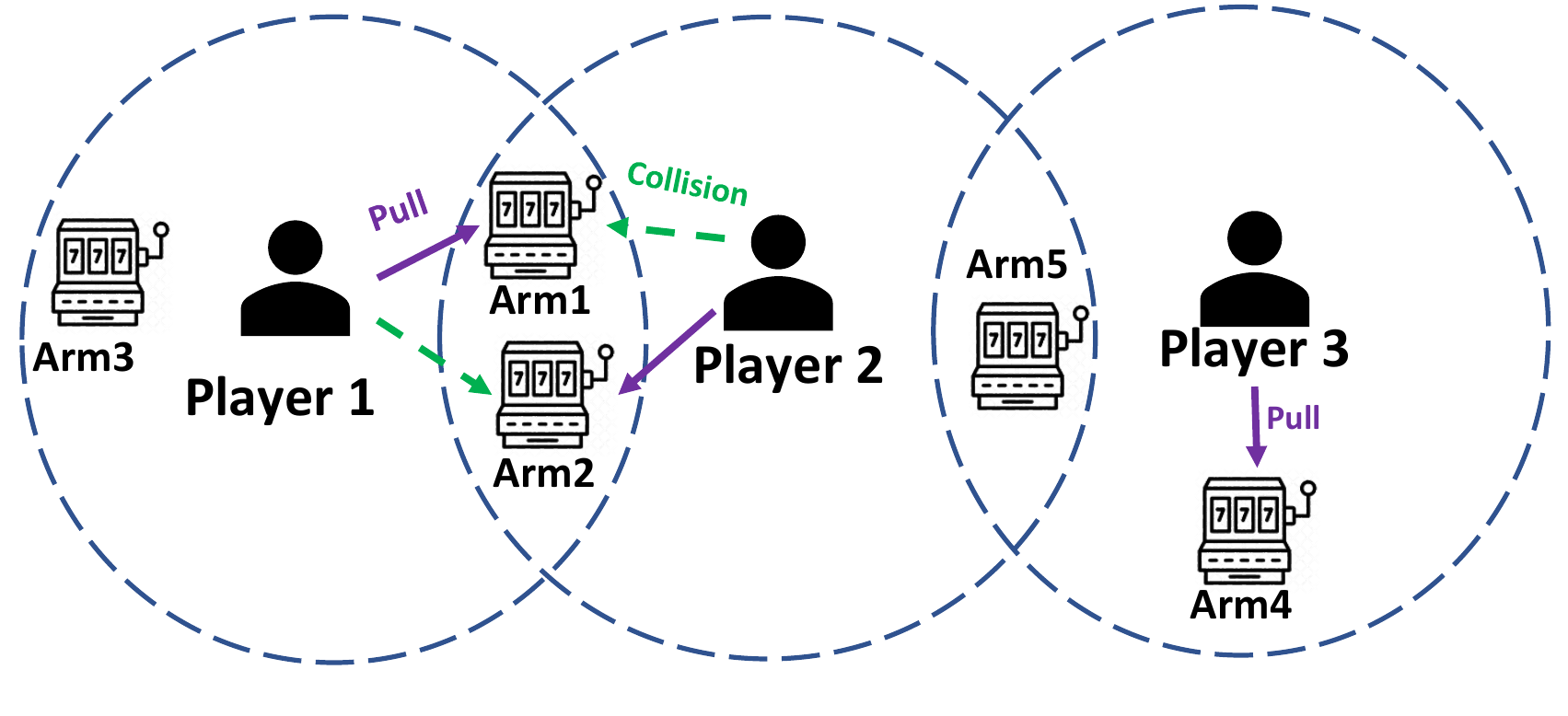}
\caption{System model for Example \ref{example1}.} 
	\label{fig:2}
    \end{figure}
\begin{example}[\match and \rank illustration]\label{example1}
Suppose there are $3$ players and $5 $ arms. At time $t$, player $1$ has the local moving arm set $\cS_1(t)=\{1,2,3\}$, player $2$ has the local moving arm set $\cS_2(t)=\{1,2,5\}$, and player $3$ has the local moving arm set $\cS_3(t)=\{4,5\}$. The system model is depicted in Figure \ref{fig:2}.  From player $2$'s perspective, it has two neighbors, i.e, $\cH_2=\{1,2,3\}$. W.l.o.g, we assume that the statistics $\bq_2(t)$ satisfies $q_{2,1}(t)\geq q_{2,2}(t)\geq q_{2,3}(t)\geq q_{2,4}(t)\geq q_{2,5}(t)$. Then, procedures of \match go as follows:  
 \begin{enumerate}
     \item [1)] player $2$ constructs the set $\cA_2=\{1, 2, 3, 4, 5\}$.
     \item [2)] Start iteration index $h=1$. player $2$ finds all players that can play arm $\cA_2^1=1$ and stores their ID in set $\cX_2^1$.  Hence $\cX_2^1=\{1,2\}$.
     \item [3)] Since $|\cX_2^1|=2\geq 1$, remove arm $1$ from set $\cA_2$ and thus $\cA_2=\{2,3,4,5\}$. player $2$ also adds arm $1$ into set $\cO_2(t)$.
     \item [4)] Now set $h=2$ and player $2$ finds players $\{1,2\}$ can play arm $\cA_2^1=2$ and sets $\cX_2^2=\{1,2\}$.
     \item [5)] Since $|\cX_2^1\cup\cX_2^2|=2\geq 2$, remove arm $2$ from $\cA_2$ and add arm $2$ into $\cO_2(t)$. Now $\cA_2=\{3,4,5\}$ and $\cO_2(t)=\{1,2\}.$
     \item [6)] Now set $h=3$ and player $2$ finds that only player $1$ can pull arm $\cA_2^1=3$ and thus sets $\cX_2^3=\{1\}$. 
     \item [7)] Since $|\cX_2^1\cup\cX_2^2\cup\cX_2^3|=2< 3$, remove arm $3$ from $\cA_2$ and keep $\cO_2(t)$ unchanged. Now $\cA_2=\{4,5\}$ and $\cO_2(t)=\{1,2\}.$
     \item [8)] Keep $h=3$ and player $2$ finds that only player $3$ can pull arm $\cA_2^1=4$ and thus sets $\cX_2^3=\{3\}$.
     \item [9)] Since $|\cX_2^1\cup\cX_2^2\cup\cX_2^3|=3\geq 3$, remove arm $4$ from $\cA_2$ and add arm $4$ into $\cO_2(t)$ unchanged. Now $\cA_2=\{5\}$ and $\cO_2(t)=\{1,2,4\}.$
     
    \item[10)] $\cS_1^*(t)=\{1,2\}$, $\cS_2^*(t)=\{1,2\}$, and $\cS_3^*=\{4\}$. Thus, there are two optimal polices for player $2$, which are $\cU_2^*(t)=\{(1,2,4), (2,1,4)\}$.
 \end{enumerate}
 Since the optimal policies indicate that player $2$ can either pull arm $1$ or arm $2$, the \rank algorithm aims to assign an unique arm for player $2$. The procedures go as follows: 
 \begin{enumerate}
     \item [1)] player $2$ constructs the set $\cI_2$ as $\{1,2\}$.
     \item[2)] Since $\cI_2=\{1,2\}\subseteq\cS_1(t)$ and $\cI_2=\{1,2\}\subseteq\cS_2(t)$, player $2$ constructs the set $\cJ_2=\{1,2\}$. 
     \item[3)] Find the rank of player $2$ in $\cJ_2$ in a decreasing order, i.e., $\beta_2=1$.
     \item[4)] Therefore, player $2$ pulls arm $\cI_2^{\beta_2}=1$.
 \end{enumerate}
 This returns the unique pulling strategy for player $1$. 
\end{example}

{
\begin{remark}
In Example \ref{example1}, the possible collision occurs between player 1 and player 2 if both players share the IDs of the available local moving arm sets under conditions that all q-statistics are correct. Specifically, player 1 and player 2 can both pull arm 1 and arm 2. 
It is impossible to avoid this collision in current time slot without adopting further actions.  This motivates us to propose the \match and \rank algorithms.  For example \ref{example1}, the \match returns all possible optimal combinations of pulling strategies at each player's perspective, which are $\{(1,2,4), (2,1,4)\}$.  The \rank outputs the unique arm each player should pull, i.e., player 1 pulls arm 2, player 2 pulls arms 1, and player 3 pulls arm 4. This avoids collisions.  
\end{remark}}


\section{The \cmabucbma Algorithm}\label{sec:commless-alg}
In this section, we consider another typical framework where players only share the information of local moving arm sets with their neighbor players. The fundamental advantage of such less information sharing is to leverage the local moving arm sets from neighbor players only to avoid collisions.  We present \cmabucbma, an algorithm similar to \cmabucber and adapted to this new setting.

\subsection{Algorithm Overview}
At each time $t$, \cmabucbma starts with an information sharing phase, which is similar to that of \cmabucber but each player $i$ only obtains the local moving arm sets  from her neighbor players $\forall j\in\cN_i$.  Since \cmabucbma does not obtain local reward estimations from neighbor players, it maintains a local empirical mean reward only based on $\hat{\mu}_{i,k}(t)$ according to
  \begin{align}\label{eq:esti_reward}
    \hat{\mu}_{i,k}(t)=\frac{\sum\limits_{\tau=1}^t\mathds{1}_{\{a_i(\tau)=k, a_j(\tau)\neq k, \forall j\neq i\}}X_{i,k}(\tau)}{I_{i,k}(t)-C_{i,k}(t)}.
\end{align}
Then \cmabucbma leverages this local estimations into the exploration phase to define an index for each arm $k$ as $g_{i,k}(t)=\hat{\mu}_{i,k}(t)+B_{i,k}(t),$  
with $B_{i,k}(t)$ being a function of $I_{i,k}(t)$ and $C_{i,k}(t)$.   Finally, \cmabucbma runs the exploitation phase in the same manner as \cmabucber besides using $g_{i,k}(t)$ as an index for arm $k$ instead of $q_{i,k}(t)$. The entire procedures are summarized in Algorithm \ref{alg:procedure1}.

\begin{algorithm}[t]
 
	\caption{\cmabucbma} 
	\label{alg:procedure1}
	\textbf{Initialize:}
	The feasible arm sets  $\{\mathcal{S}_i(1), \forall i\in \mathcal{N}\}$; 
	the sample mean available at player $i$ $\{\hat{\mu}_{i,k}(1)=0, \forall k\}$, and the statistics $\{g_{i,k}(1)=\infty,\forall k\}$;
	the number of pulls $\{I_{i,k}(1)=0, \forall  k\}$ and the number of collisions $\{C_{i,k}(1)=0, \forall k\}$.
	\begin{algorithmic}[1]
		\FOR{$t=1,1,...,T$}

        \STATE Transmit  $\mathcal{S}_i(t)$ to its neighbors and receive  $\mathcal{S}_j(t)$ from all the neighbor players $j\in\mathcal{N}_i$.\\
        
        \STATE  Solving the matching problem (defined in (7) of the main paper) by  \match and selects the arm indicated as $\ba_i^*(1)$ by \rank. \\

        \STATE Updates $I_{i,k}(t)$ and $C_{i,k}(t)$;  $\hat{\mu}_{i,k}(t+1)$  and $g_{i,k}(t+1)$.

		\ENDFOR
	\end{algorithmic}
\end{algorithm}

\subsection{Regret Analysis}\label{sec:commless-alg-regret}

\textbf{Theorem 2.} 
\emph{For $B_{i,k}(t)=\sqrt{\frac{3\log t}{2V_{i,k}(t)}}$, the regret of \cmabucbma satisfies} 
\begin{align*}
    R(T)&\leq \mu_1N\Bigg(\sum_{j\in\cK_{-b}}\sum_{k\in\cK\setminus\{j\}} \frac{6\log T}{(\mu_k-\mu_j)^2}\\
    &+\sum_{k=1}^N\sum_{m=k+1}^K \frac{6\log T}{(\mu_k-\mu_m)^2}+\frac{\pi^2}{3}K(K+N)\Bigg). 
\end{align*}

\begin{remark}
Similar to Theorem 1, the regret of \cmabucbma is also incurred by three terms and sub-logarithmic in time $T$.
The first two terms scale with $\cO(NK^2\log T)$.  From  Theorem 1, it is clear that \cmabucber attains an improved regret bound with a factor of $\mathcal{O}(N)$ compared to that of \cmabucbma.  This is intuitive  since player $i$ in \cmabucber also receives the reward estimation from her neighbors $\cN_i$ at each time, where $|\cN_i|<N.$
The estimated reward sharing in \cmabucber can be approximately regarded as a means to improve the exploration efficiency by a factor of $\mathcal{O}(N)$, i.e., a $\cO(N)$ decrease for the number of time steps needed to obtain the correct statistics of arms.  The number of communication bits is upper bounded by ${\mathcal{O}}(N^2KT)$.  
It is not surprising that 
the dynamic subset of arms regularly brings external randomness as mentioned in Remark~\ref{remark:regret-definition}, resulting in the multiplicative pre-factor that goes with the time-horizon to be $K^2N$ in the regret compared to $KN$ in \cite{wang2020optimal}. 

\end{remark}


\section{Summary of Notations}

We list all notations used in this paper in Table~\ref{tbl:notation}.

\begin{table*}
\caption{Summary of Notations}
\begin{center}
\begin{tabular}{ | m{6em} | m{10cm} | } 
 \hline
 Notations & Definitions\\ 
 \hline
$\cN; N$ &Set of players; number of players \\ 
$\cK; K$ &Set of arms; number of arms\\
$T$ & Time Horizon\\
$\mu_i$& Mean reward of arm $i$\\
$X_{i,k}(t)$& Reward of arm $k$ by player $i$ at time $t$\\
$\hat{\mu}_{i,k}(t)$& Empirical estimation of reward of arm $k$ at player $i$ till $t$\\
$\mathcal{S}_{i}(t)$& The set of available arms by player $i$ at time $t$\\
$\cN_i$& Set of neighbor players of player $i$\\
$I_{i,k}(t); C_{i,k}(t)$& Number of pulls and number of collisions for arm $k$ at player $i$ till $t$;\\
$B_{i,k}(t)$& Upper confidence bound for arm $k$ at player $i$ at time $t$ \\
$g_{i,k}(t)$& Index for exploration and exploitation of \cmabucbma
\\
$q_{i,k}(t)$& Index for exploration and exploitation of \cmabucber
\\
$\tilde{r}_{i,k}(t)$& Estimation of reward of arm $k$ at player $i$ till $t$ for \cmabucber\\
$V_{i,j}(t)$&  Number of times that arm $j$ is only pulled by player $i$ by time $t$.\\
$\cK_b$ &Set containing arms with the largest $N$ mean reward\\
$\cK_{-b}$& Set of the $K-N$ remaining arms\\
$C(T)$ & Number of collisions faced by players by pulling arms in $\cK_b$ by time $T$\\
 \hline
\end{tabular}
\end{center}
\label{tbl:notation}
\end{table*}


\section{Proofs of Main Results}\label{sec:proof-app}

In this section, we provide the proofs of theoretical results presented in the paper. 

\noindent\textbf{Notations.}
Let $N$ be the total number of players, $K$ be the number of arms, and $T$ be the total time horizon.  
Let $V_{i,j}(t)$ be the number of times that arm $j$ is only pulled by player $i$ by time $t$, and denote $V_j(t):=\sum_{i=1}^N V_{i,j}(t)$.  
Furthermore, we define $I_j(t):=\sum_{i=1}^N I_{i,j}(t)$, where $I_{i,j}(t)$ is the number of times player $i$ pulling arm $j$ by time $t$.    We denote $\cK_b$ as the set containing arms with the largest $N$ mean rewards, i.e.,  $\cK_b:=\{\mu_1, \mu_2, \ldots, \mu_N\}.$ We let $\cK_{-b}=\mathcal{K}\setminus \cK_b$ contain the remaining arms.  Finally, let $C(T):=\sum_{j\in \cK_b}C_j(T)=\sum_{j\in \cK_b}I_j(T)-V_{j}(T)$ be the number of collisions faced by players by pulling arms in $\cK_b$ by time $T$.

\subsection{Proof of Theorem~\ref{thm:A2}}
The following lemma upper bounds the regret by $C(T)$ and $I_j(T)$.
 
\begin{lemma}
\label{prop:ub}
The regret of \cmabucber is upper-bounded by
\begin{align}\nonumber
  R(T)\leq\mu_1\Bigg(\sum_{j\in\cK_b}\mathbb{E}[C_j(T)]+ \sum_{j\in \cK_{-b}}\mathbb{E}[I_{j}(T)]\Bigg),
\end{align}
where the first and second terms respectively correspond to the regret incurred by playing each suboptimal arms and the collisions occurred on playing the best $N$ arms in $\cK_b$. 
\end{lemma}

\begin{remark}
We adopt a similar decomposition of regret to number of collisions and number of pulls as \cite{anandkumar2011distributed}. To make this paper self-contained, we provide the detailed proof as below.
\end{remark}
\begin{proof}
It is easy to see that $\sum_{j=1}^K I_j(T)=TN$.
Based on the definition of regret,
we have
\begin{align}
 R(T)&=\sum_{t=1}^{T}\sum_{n=1}^{N}\mu_{a^*_n(t)}- \sum_{j=1}^K\mu_j\mathbb{E}[V_{j}(T)]\nonumber\displaybreak[0]\\
    &\overset{(a1)}{\leq} \sum_{j\in \cK_b}\mu_j\Big(T-\mathbb{E}[V_{j}(T)]\Big)\nonumber\displaybreak[1]\\
    &\overset{(a2)}{\leq} \mu_1\left(TN- \sum_{j\in \cK_b}\mathbb{E}[V_{j}(T)]\right)\nonumber\displaybreak[2]\\
    &\overset{(a3)}{\leq}\mu_1\left(\sum_{j\in\cK_b}\mathbb{E}[C_j(T)]+ \sum_{j\in\cK_{-b}}\mathbb{E}[I_{j}(T)]\right),\label{eq:upperbound}
\end{align}
where $(a1)$ follows from the fact that the highest sum reward at each time cannot exceed the sum reward of the $N$ best arms, i.e., $\sum_{n=1}^{N}\mu_{a^*_n(t)}\leq \sum_{j\in \cK_b}\mu_j$. $(a2)$ is due to $T-V_j(T)\geq 0$ and $\mu_1$ is the largest reward. Note that $TN=\sum_{j\in \cK_b}\mathbb{E}[I_{j}(T)]+\sum_{j\in \cK_{-b}}\mathbb{E}[I_{j}(T)]$, and  $(a3)$ follows from the definition of $C_j(T)$. 
\end{proof}

Our key innovation to prove Theorem~\ref{thm:A2} is to bound the number of collisions  $C_j(T), \forall j\in\cK_b$ and the time spent on each arm $I_j(T), \forall j\in\cK_{-b}$.
Under the communication through collision framework,
 we first need to characterize the property of random variables $\tilde{r}_{i,k}(t), \forall i\in\cN, \forall k\in\cK$ to derive the regret.

\begin{lemma}
For any $i\in\cN, k\in\cK$ and time $t\geq 0$, $\tilde{r}_{i,k}(t)$ is a  random variable with mean $\mu_k$.
\end{lemma}
\begin{proof}
We prove this by induction. At $t=1$, we have $\mathbb{E}[\tilde{r}_{i,k}(1)]=\mathbb{E}[\hat{\mu}_{i,k}(1)]=\mu_k, \forall i\in\cN$. Assume that $\mathbb{E}[\tilde{r}_{i,k}(t)]=\mu_k, \forall i\in\cN$. We need to show that
\begin{align*}
\mathbb{E}[\tilde{r}_{i,k}({t+1})]&=\sum_{j\in \mathcal{N}_j} \mathbb{E}[\tilde{r}_{j,k}(t)]P_{i,j}+ \mathbb{E}[\hat{\mu}_{i,k}(t+1)]-\mathbb{E}[\hat{\mu}_{i,k}(t)]  
\\&=\sum_{j\in \mathcal{N}_j}P_{i,j}\mu_k=\mu_k.
\end{align*}
The equality holds due to $\mathbb{E}[\hat{\mu}_{i,k}(t+1)]-\mathbb{E}[\hat{\mu}_{i,k}(t)]=0$ and $\sum_{j\in \mathcal{N}_j}P_{i,j}=1.$
\end{proof}

\begin{lemma}\label{lemma:variance}
For any $i\in\cN, k\in\cK$ and time $t\geq 0$, $\tilde{r}_{i,k}(t)$ is a sub-Gaussian random variable, and 
the optimal variance proxy of $\tilde{r}_{i,k}(t)$ is no larger than $\frac{3}{8|\mathcal{N}_i|V_{i,k}(t)}$ if $V_{i,k}(t)\geq L$ with 
$$
  L=\min_t  3(1-\beta^N)^{t/24N(1+\beta^{-N})}\leq \frac{(1-\beta^N)}{48N(1+\beta^{-N})t},
$$
where $\beta$ is the smallest positive value of all consensus matrices, i.e., $\beta=\arg\min P_{i,j}$ with $P_{i,j}>0, \forall i,j.$
\end{lemma}
\begin{proof}
 $\tilde{r}_{i,k}(t)$ is sub-Gaussian directly follows Lemma \ref{lemma:bound_G}, since it is bounded almost surely and linear combination of $X_{j,k}(\tau), \forall j\in[N], \tau\in\{1,2,\ldots,t\}$. 
 Define $\tilde{\br}_k:=[\tilde{r}_{1,k},  \ldots, \tilde{r}_{N,k}]^\intercal$ and  $\hat{\pmb{\mu}}_k:=[\hat{\mu}_{1,k},  \ldots, \hat{\mu}_{N,k}]^\intercal$as the vector stack of $\tilde{r}_{i,k}$ and $\hat{\mu}_{i,k}$, respectively. { Then, we have
\begin{align}
    \tilde{\br}_k(t)=\bP\tilde{\br}_k(t-1)+(\hat{\pmb{\mu}}_k(t)-\hat{\pmb{\mu}}_k(t-1)).
    \label{eq:matrix_form}
\end{align}
Based on \eqref{eq:matrix_form}, we have the following expression with respect to the vector $\tilde{\br}_k(t)$ as
\begin{align}
    \tilde{\br}_{k}(t)&=\bP \tilde{\br}_{k}(t-1)+\hat{\pmb{\mu}}_k(t)-\hat{\pmb{\mu}}_k(t-1)\nonumber\displaybreak[0]\\
    &=\bP^t\tilde{\br}_{k}(0)+\sum_{s=0}^{t-1}\bP^s(\hat{\pmb{\mu}}_k(t-s)-\hat{\pmb{\mu}}_k(t-s-1))\nonumber\displaybreak[1]\\
    &=\sum_{s=0}^{t-1}\bP^s\hat{\pmb{\mu}}_k(t-s)-\sum_{s=0}^{t-1}\bP^s\hat{\pmb{\mu}}_k(t-s-1),
\end{align}
where the second inequality is due to the fact that $\tilde{\br}_k(0)=0$,} Thus, we have the following expression for arm $k$ at player $i$
\begin{align}\label{eq:r_tilde_esti}
    \tilde{r}_{i,k}(t)=\sum_{j}\sum_{s=0}^{t-1}{ [\bP^s]}_{ij}(\hat{{\mu}}_{j,k}(t-s)-\hat{{\mu}}_{j,k}(t-s-1)),
\end{align}
with$[\bP^s]_{ij}$ being the $i$-th row $j$-th column element of $\bP^s$. 
Note that $\hat{\mu}_{j,k}(t)-\hat{\mu}_{j,k}(t-1)$ in \eqref{eq:r_tilde_esti} is zero when arm $k$ is not pulled by player $j$ or  collision occurs when arm $k$ is pulled by player $j$ at time $t$. Denote $\tau_{i,1}, \tau_{i,2}, \ldots, \tau_{i, V_{i,k}(t)}$ as the time instance at which player $i$ is the only player that pulls arm $k$, then we have 
\begin{align}
    \tilde{r}_{i,k}(t)&=\sum_{j}\sum_{s=1}^{V_{j,k}(t)}[\bP^{t-\tau_{j,s}}]_{ij}(\hat{{\mu}}_{j,k}(\tau_{j,s})-\hat{{\mu}}_{j,k}(\tau_{j,s}-1))\nonumber\displaybreak[0]\\
    &=\sum_j\Bigg(\sum_{s=1}^{V_{j,k}(t)}[\bP^{t-\tau_{j,s}}]_{ij}\hat{\mu}_{j,k}(\tau_{j,s})\nonumber\\
    &\qquad\qquad-\sum_{s=1}^{V_{j,k}(t)}[\bP^{t-\tau_{j,s}}]_{ij}\hat{\mu}_{j,k}(\tau_{j,s}-1)\Bigg)\nonumber\displaybreak[1]\\
    &=\sum_j\Bigg(\sum_{s=1}^{V_{j,k}(t)-1}[\bP^{t-\tau_{j,s}}]_{ij}\hat{\mu}_{j,k}(\tau_{j,s})\nonumber\\
    &\qquad\qquad-\sum_{s=1}^{V_{j,k}(t)-1}[\bP^{t-\tau_{j,s+1}}]_{ij}\hat{\mu}_{j,k}(\tau_{j,s+1}-1)\nonumber\displaybreak[2]\\
    &\qquad\qquad+[\bP^{t-V_{j,k}(t)}]_{ij}\hat{\mu}_{j,k}(\tau_{j,V_{j,k}(t)})\nonumber\\
    &\qquad\qquad-[\bP^{t-\tau_{j,1}}]_{ij}\hat{\mu}_{j,k}(\tau_{j,1}-1)\Bigg)\nonumber\displaybreak[3]\\
  &\overset{(e)}{=}\sum_j\Bigg(\sum_{s=1}^{V_{j,k}(t)-1}[\bP^{t-\tau_{j,s}}-\bP^{t-\tau_{j,s+1}}]_{ij}\hat{\mu}_{j,k}(\tau_{j,s})\nonumber\\
  &\qquad\qquad+[\bP^{t-\tau_{j,V_{j,k}(t)}}]_{ij}\hat{\mu}_{j,k}(\tau_{j,V_{j,k}(t)})\Bigg),
\end{align}
where $(e)$ holds due to the fact that $\hat{\mu}_{j,k}(\tau_{j,s})=\hat{\mu}_{j,k}(\tau_{j,s+1}-1)$ and $\hat{\mu}_{j,k}(\tau_{j,1}-1)=0$ according to the definition.

Let $c_{i,k,j}^{\tau}(t)$ be the coefficient of $X_{j,k}(\tau)$ in $\tilde{r}_{i,k}(t)$, we have
\begin{align*}
    \sigma_{i,k}^2=\frac{1}{4}\sum_{j=1}^{N}\sum_{h=1}^{V_{j,k}(t)}|c_{i,k,j}^{(\tau_{j,h})}(t)|^2.
\end{align*}
We also can denote the coefficient as 
\begin{align*}
    c_{i,k,j}^{\tau_{j,\ell}}=\Bigg[\sum_{h=\ell}^{V_{j,k}(t)-1}\frac{\bP^{t-\tau_{j,h}}-\bP^{t-\tau_{j,h+1}}}{h}+\frac{\bP^{t-\tau_{j,V_{j,k}(t)}}}{V_{j,k}(t)}\Bigg]_{ij}.
\end{align*}
Note that $c_{i,k,j}^{\tau_{j,\ell}}(t)$ is non-increasing in $\tau_{j,\ell}$, thus we can bound the first term $c_{i,k,j}^{\tau_{j,1}}(t)$.
We first rewrite $c_{i,k,j}^{\tau_{j,1}}(t)$ as
\begin{align*}
   c_{i,k,j}^{\tau_{j,1}}(t)=\Bigg[\bP^{t-\tau_{j,1}}-\sum_{h=2}^{V_{j,k}(t)}\frac{\bP^{t-\tau_{j,h}}}{(h-1)h}\Bigg]_{ij}.
\end{align*}
According to Lemma \ref{themorem_convergence_Phi} and Lemma \ref{lemma_bound_Phi}, $\Big|[\bP^t]_{ij}-[\bP_\infty]_{ij}\Big|\leq 2 {(1+\beta^{-N})}(1-\beta^{N})^{\frac{t}{N}-1}$. Hence
we have
\begin{align*}
  |c_{i,k,j}^{\tau_{j,1}}(t)|&\leq \frac{[\bP_\infty]_{ij}}{V_{i,k(t)}}+2 {(1+\beta^{-N})}(1-\beta^{N})^{\frac{t-\tau_{j,1}}{N}-1}\\
  &+\sum_{h=2}^{V_{j,k}(t)}\frac{2 {(1+\beta^{-N})}(1-\beta^{N})^{\frac{t-\tau_{j,h}}{N}-1}}{(h-1)h}.
\end{align*}
According to Lemma 7 in \cite{ZL21}, let $L$ be the smallest time such that  
\begin{align*}
    3(1-\beta^N)^{L/24N(1+\beta^{-N})}\leq \frac{(1-\beta^N)}{48N(1+\beta^{-N})L}.
\end{align*}
Then if $V_{j,k}(t)\geq L$, we have 
$
    |c_{i,k,j}^{\tau_{j,1}}(t)|\leq \frac{[P_\infty]_{ij}}{V_{j,k(t)}}+\frac{1}{8NV_{j,k}(t)}.
$
Hence we obtain
\begin{align}
     \sigma_{i,k}^2&=\frac{1}{4}\sum_{j=1}^{N}\sum_{h=1}^{V_{j,k}(t)}|c_{i,k,j}^{(\tau_{j,h})}(t)|^2\nonumber\displaybreak[0]\\
     &\leq \frac{1}{4V_{i,k}(t)}\Bigg(\sum_{j=1}^N[\bP_\infty]^2_{ij}+\frac{1}{4N}\sum_{j=1}^{N}[\bP_\infty]_{ij}+\frac{1}{64N}\Bigg)\nonumber\displaybreak[1]\\
     &\leq \frac{1}{4V_{i,k}(t)}\Bigg(\frac{1}{N}+\frac{1}{4N}+\frac{1}{64N}\Bigg)\nonumber\displaybreak[2]\\
     &\leq \frac{3}{8NV_{i,k}(t)}.
\end{align}
This completes the proof.
\end{proof}

\begin{lemma}\label{lem:5}
The total time spent by any player $i$ on the worst $K-N$ arms in \cmabucber is given by
\begin{align}
   \mathbb{E}[I_{i,j}(T)]\leq \sum_{k\in\cK\setminus j} \max\left\{\frac{6\log T}{N(\mu_k-\mu_j)^2},L\right\}+\frac{\pi^2}{3}K. 
\end{align}
\end{lemma}

\begin{proof}
Denote the desired arm to be pulled by player $i$ at time $t$ as $\pi_i(t)$ based on $\ba_i^{*,1}(t)$. Then we have 
\begin{align*}
    \mathds{1}_{\{\text{player $i$ pull arm $j$ at $t$}\}}=\mathds{1}_{\{\pi_i(t)=j\}}.
\end{align*}
Define the following event that there exists at least one arm such that the true mean is outside of the confidence interval, i.e., 
\begin{align}\label{eq:A1_bad}
   \cB_i(t)&=\{ \exists k\in\cS_i(t):|\mu_k-\tilde{r}_{i,k}(t)|\geq B_{i,k}(t)\},
\end{align}
which is a rare event that happens with small probability. We also define the complementary event as
\begin{align}\label{eq:A1_good}
    \bar{\cB}_i(t)&=\{ \forall k\in\cS_i(t):|\mu_k-\tilde{r}_{i,k}|\leq B_{i,k}(t)\}.
\end{align}
Hence, we can decompose the event that player $i$ pulls arm $j$ at round $t$ as two disjoint parts as
\begin{align}
\label{eq:decomp_pull}
    \mathds{1}_{\{\text{player $i$ pull arm $j$ at $t$}\}}=\mathds{1}_{\{\pi_i(t)=j\}}\mathds{1}_{\{\cB_i(t)\}}+\mathds{1}_{\{\pi_i(t)=j\}}\mathds{1}_{\{\bar{\cB}_i(t)\}}.
\end{align}
Taking the expectation of both sides of \eqref{eq:decomp_pull} and summing up it to $T$, we have
\begin{align}
    &\mathbb{E}[I_{i,j}(T)]=\sum_{t=1}^T { \mathbb{P}}[\text{player $i$ pulls arm $j$ at time $t$}]\nonumber\displaybreak[0]\\
    &=\sum_{t=1}^T\mathbb{E}[\mathds{1}_{\{\pi_i(t)=j\}}\mathds{1}_{\{\cB_i(t)\}}]+\sum_{t=1}^T\mathbb{E}[\mathds{1}_{\{\pi_i(t)=j\}}\mathds{1}_{\{\bar{\cB}_i(t)\}}]\nonumber\displaybreak[1]\\
    &\leq \underset{Term_1}{\underbrace{\sum_{t=1}^T \mathbb{P}[\cB_i(t)]}}+\underset{Term_2}{\underbrace{\sum_{t=1}^T \mathbb{P}[\{\pi_i(t)=j\}\cap \bar{\cB}_i(t)]}}.\nonumber
\end{align} 
According to Lemma \ref{lemma:variance} and Lemma \ref{lem:subG_inequality}, we have
\begin{align}\label{eq:term1}
{{\sum_{t=1}^T \mathbb{P}[\cB_i(t)]}}&\leq \sum_{t=1}^T\sum_{k\in\cS_i(t)} \mathbb{P}(|\mu_k-\tilde{r}_{i,k}(t)|\geq B_{i,k}(t)) \nonumber\\
&\leq \sum_{t=1}^T\sum_{k\in\cS_i(t)} \frac{2}{t^2}
\leq \frac{\pi^2}{3}K.
\end{align}

Next, we need to bound $Term_2$. 
First, we define the following events
\begin{align*}
    \mathcal{\cE}^1(t)&:=\bigcup_{k\in \cK_b} q_{i,k}(t)\leq q_{i,j}(t), \\
    \mathcal{E}^2(t)&:=\{k\notin \mathcal{S}_i(t)|k\in \cK_b\},\\
    \mathcal{E}^3(t)&:=\bigcup_{k\in \cK_{-b}\setminus j} q_{i,k}(t)\leq q_{i,j}(t).
\end{align*}
Then, we have for any arm $j\in \cK_{-b}$, the player $i$ pulls arm $j$ under the event $\bar{\cB}_i(t)$ with probability
\begin{align*}
   & \mathbb{P}[\{\pi_i(t)=j\}]\leq\mathbb{P}\Big[\Big((q_{i,N}(t)\leq q_{i,j}(t)) \cup \mathcal{E}^2(t)\Big)\cap \mathcal{E}^3(t)\Big]\\
    &\leq\mathbb{P}\big[(q_{i,N}(t)\leq q_{i,j}(t))\big]+\mathbb{P}\big[ \mathcal{E}^2(t)\cap \mathcal{E}^3(t)\big]\\
    &= \mathbb{P}[\mathcal{E}^1(t)\cap (q_{i,N}(t)\leq q_{i,j}(t))]\\
&\qquad    +\mathbb{P}[\bar{\mathcal{E}^1}(t)\cap (q_{i,N}(t)\leq q_{i,j}(t))]+\mathbb{P}[ \mathcal{E}^2(t)\cap \mathcal{E}^3(t)]\\
    &\overset{(c1)}{=} \mathbb{P}[\mathcal{E}^1(t)\cap (q_{i,N}(t)\leq q_{i,j}(t))]+\mathbb{P}[ \mathcal{E}^2(t)\cap \mathcal{E}^3(t)]\\
    &=\sum_{k\in \cK_b}\mathbb{P}[q_{i,k}(t)\leq q_{i,j}(t)]+\mathbb{P}[ \mathcal{E}^2(t)\cap \mathcal{E}^3(t)]\\
    &\leq \sum_{k\in \cK\setminus j}\mathbb{P}[q_{i,k}(t)\leq q_{i,j}(t)],
\end{align*}
where $(c1)$ holds due to the fact that 
$\mathbb{P}[\bar{\mathcal{E}^1}(t)\cap (q_{i,N}(t)\leq q_{i,j}(t))]=0.$
Hence, we bound $Term2$ as 
\begin{align}
 Term_2&\leq  \mathbb{E}\left[\sum_{t=1}^T\sum_{k\in \cK\setminus j}\mathds{1}[q_{i,k}(t)\leq q_{i,j}(t)]\right]\nonumber \\
& =\mathbb{E} \left[\sum_{k\in \cK\setminus j}\sum_{t=1}^T\mathds{1}[q_{i,k}(t)\leq q_{i,j}(t)]\right].
 \label{eq:term2}
\end{align}
For the second term, we have
\begin{align}\nonumber
 {{\sum_{t=1}^T \mathbb{P}[\{\pi_i(t)=j\}\cap \bar{\cB}_i(t)]}}\leq  \mathbb{E} \left[\sum_{k\in \cK\setminus j}\sum_{t=1}^T\mathds{1}[q_{i,k}(t)\leq q_{i,j}(t)]\right].
 \label{eq:term2}
\end{align}
For arbitrary $k$, the indicator function $\mathds{1}[q_{i,k}(t)\leq q_{i,j}(t)]=1$ equals to the condition that $\mu_k\leq \mu_{j}+2B_{i,j}(t)$.
It is easy to verify that if $I_{i,j}(t)\geq \frac{6\log T}{N(\mu_k-\mu_j)^2}$, the condition will not hold. Under the condition $I_{i,j}(t)\geq L$, it gives rise to the bound 
\begin{align*}
    {{\sum_{t=1}^T \mathbb{P}[\{\pi_i(t)=j\}\cap \bar{\cB}_i(t)]}}\leq \sum_{k\in\cK\setminus j} \max\left\{\frac{6\log T}{N(\mu_k-\mu_j)^2},L\right\}. 
\end{align*}
Therefore, we have
\begin{align*}
    \mathbb{E}[I_{i,j}(T)]\leq \sum_{k\in\cK\setminus j} \max\left\{\frac{6\log T}{N(\mu_k-\mu_j)^2},L\right\}+\frac{\pi^2}{3}K. 
\end{align*}
\end{proof}

\begin{lemma}\label{lem:A2_2}
For \cmabucber, the number of collisions faced by the player in the $N$-best arms in $T$ can be upper bounded as
\begin{align}
     \mathbb{E}[C(T)]\leq \sum_{k=1}^N\sum_{m=k+1}^K \max\left\{\frac{6\log T}{(\mu_k-\mu_m)^2},L\right\}+\frac{2\pi^2}{3}KN^2.
\end{align}
\end{lemma}
\begin{proof}
Under perfect knowledge of each arm, along with the information of local moving arm sets from neighbor players, each player can construct an optimal policy based on its local information without causing any collision. The collision may only occurs when the $q$-statistics are not correct. 
 For player $i$,  define the following event
\begin{align*}
    \mathcal{W}_i(t):=\{\exists (m,k), m\in\cK, k\in \cK_b| q_{i,m}(t)> q_{i,k}(t), m> k\},
\end{align*}
where $\cW_i(t)$ is non-empty denoting the event that the $\bg$-statistics of player $i$ at time $t$ is not accurate. 
Hence, $C(T)$ is bounded by
\begin{align}
    \mathbb{E}[C(T)]&\leq \mathbb{E}\left[\sum_{t=1}^T\sum_{i=1}^N\mathds{1}_{\{\mathcal{W}_i(t)\}}\right]\nonumber\\
&    \leq \sum_{t=1}^T\sum_{i=1}^N\sum_{k=1}^N\sum_{m=k+1}^K\mathbb{E}[\mathds{1}_{\{q_{i,k}(t)<q_{i,m}(t)\}}].\label{eq:A1_C}
\end{align}
The goal is to bound $\sum_{t=1}^T\mathbb{E}[\mathds{1}_{\{q_{i,k}(t)<q_{i,m}(t)\}}]$ for any $k\in\cM_b, m\in \cM$. Similar to the proof of Lemma \ref{lem:5}, we decompose this event into two distinct parts as
\begin{align*}
 &\sum_{t=1}^T\mathbb{E}[\mathds{1}_{\{q_{i,k}(t)<q_{i,m}(t)\}}]\\
 &=\sum_{t=1}^T\mathbb{E}[\mathds{1}_{\{q_{i,k}(t)<q_{i,m}(t)\}}\cdot\mathds{1}_{\{\cD_i(t)\}}] \\
 &\qquad+ \sum_{t=1}^T\mathbb{E}[\mathds{1}_{\{q_{i,k}(t)<q_{i,m}(t)\}}\cdot\mathds{1}_{\{\bar{\cD}_i(t)\}}] \nonumber\displaybreak[0]\\
&\leq\underset{Term_3}{\underbrace{\sum_{t=1}^T \mathbb{P}[\cD_i(t)]}}+\underset{Term_4}{\underbrace{\sum_{t=1}^T \mathbb{P}[\{q_{i,k}(t)<q_{i,m}(t)\}\cap \bar{\cD}_i(t)]}},
\end{align*}
where $\cD_i(t)$ and $\bar{\cD}_i(t)$ are defined as
\begin{align*}
     \cD_i(t)=\{|\mu_k-\tilde{r}_{i,k}|\geq B_{i,k}(t)\}\cup\{|\mu_m-\tilde{r}_{i,m}|\geq B_{i,m}(t)\},\\
    \bar{ \cD}_i(t)=\{|\mu_k-\tilde{r}_{i,k}|\leq B_{i,k}(t)\}\cap\{|\mu_m-\tilde{r}_{i,m}|\leq B_{i,m}(t)\}.
\end{align*}
Similar to \eqref{eq:term1}, $Term_3\leq \frac{2\pi^2}{3}$. To bound $Term_4$, the key is to bound $\sum_{t=1}^T \mathbb{P}[\{q_{i,k}(t)<q_{i,m}(t)\}]$ under the good event $\bar{\cD}_i(t)$, which is given
\begin{align*}
    Term4&\leq \sum_{t=1}^T \mathbb{P}[\{q_{i,k}(t)<q_{i,m}(t)\}]\\
&    \overset{(d1)}{=}\sum_{t=1}^T \mathbb{P}[\{\mu_k<\mu_m+2B_{i,m}(t)\}]
    \overset{(d2)}\leq \frac{6\log T}{N(\mu_k-\mu_m)^2},
\end{align*}
where $(d1)$ is due to the fact that $  \mu_k\leq q_{i,k}(t)\leq q_{i,j}(t)\leq \mu_j+2B_{i,m}(t)$ holds under the good event, and $(d2)$ is due to the fact that if $ t\geq \frac{6\log T}{N(\mu_k-\mu_m)^2}$ the condition in $(d1)$ will no longer hold. 

Under the condition $I_{i,j}(t)\geq L$,  substituting $Term_3$ and $Term_4$ into \eqref{eq:A1_C} yields the desired result 
\begin{align*}
    \mathbb{E}[C(T)]\leq \sum_{k=1}^N\sum_{m=k+1}^K \max\left\{\frac{6\log T}{(\mu_k-\mu_m)^2}, L\right\}+\frac{2\pi^2}{3}KN^2. 
\end{align*}
\end{proof}
Combining Lemma \ref{lem:5} and Lemma \ref{lem:A2_2} leads to the desired results in Theorem~\ref{thm:A2}.

\subsection{Proof of Theorem 2}

\begin{lemma}\label{lemma:A1-1}
{The total time spent by any player $i$ on the worst $K-N$ arms in $\cK_{-b}$ in \cmabucbma is given by
\begin{align}
     \mathbb{E}[I_{i,j}(T)]\leq \sum_{k\in\cK\setminus j} \frac{12\log T}{(\mu_k-\mu_j)^2}+\frac{\pi^2}{3}K, ~\forall j\in \cK_{-b}.
\end{align}}
\end{lemma}

\begin{proof}
Following similar procedures in Lemma \ref{lem:5},
we define the following event that there exists at least one arm such that the true mean is outside of the confidence interval, i.e., 
\begin{align}\label{eq:A1_bad}
    \cB_i^\prime(t)=\{ \exists k\in\cS_i(t):|\mu_k-\hat{\mu}_{i,k}(t)|\geq B_{i,k}(t)\},
\end{align}
which is a rare event that happens with small probability. We also define the complementary event as
\begin{align}\label{eq:A1_good}
    \bar{\cB}^\prime_i(t)=\{ \forall k\in\cS_i(t):|\mu_k-\hat{\mu}_{i,k}|\leq B_{i,k}(t)\}.
\end{align}
Then, we have
\begin{align}
    \mathbb{E}[I_{i,j}(T)]\leq \underset{Term_1}{\underbrace{\sum_{t=1}^T \mathbb{P}[\cB^\prime_i(t)]}}+\underset{Term_2}{\underbrace{\sum_{t=1}^T \mathbb{P}[\{\pi_i(t)=j\}\cap \bar{\cB}^\prime_i(t)]}}.
    \label{eq:I_decompose}
\end{align}
We first bound $Term_1$. According to Chernoff-Hoeffding inequality \cite{hoeffding1994probability}, we have
\begin{align}\nonumber
    \mathbb{P}\left(|\mu_k-\hat{\mu}_{i,k}(t)|\geq B_{i,k}(t)\right)\leq \frac{2}{t^2}.
\end{align}
Hence, we have
\begin{align}
\label{eq:term111}
Term_1&\leq \sum_{t=1}^T\sum_{k\in\cS_i(t)}  \mathbb{P}(|\mu_k-\hat{\mu}_{i,k}(t)|\geq B(t)) \nonumber\\&
\leq \sum_{t=1}^T\sum_{k\in\cS_i(t)} \frac{2}{t^2}
\leq \frac{\pi^2}{3}K.
\end{align}
The last inequality holds due to: 1) $\sum_{t=1}^T \frac{2}{t^2}\leq \sum_{t=1}^\infty \frac{2}{t^2}=\frac{\pi^2}{3}$ and 2) $|\cS_i(t)|\leq K$.

Hence, we bound $Term2$ as 
\begin{align}
 Term_2&\leq  \mathbb{E} \left[\sum_{k\in \cK\setminus j}\sum_{t=1}^T\mathds{1}[g_{i,k}(t)\leq g_{i,j}(t)]\right].
 \label{eq:term2}
\end{align}
For arbitrary $k$, the indicator function $\mathds{1}[g_{i,k}(t)\leq g_{i,j}(t)]=1$ equals to the condition that $\mu_k\leq \mu_{j}+2B_{i,j}(t)$, which is due to 
    $\mu_k\leq g_{i,k}(t)\leq g_{i,j}(t)\leq \mu_j+2B_{i,j}(t).$
It is easy to verify that if $I_{i,j}(t)\geq \frac{6\log T}{(\mu_k-\mu_j)^2}$, the condition will not hold. This gives rise to the bound that 
\begin{align*}
    Term_2\leq \sum_{k\in\cK\setminus j}\frac{6\log T}{(\mu_k-\mu_j)^2}.
\end{align*}
{Substituting $Term_1$ and $Term_2$ back into \eqref{eq:I_decompose}, we have 
\begin{align*}
    \mathbb{E}[I_{i,j}(T)]\leq \sum_{k\in\cK\setminus j} \frac{6\log T}{(\mu_k-\mu_j)^2}+\frac{\pi^2}{3}K. 
\end{align*}}
\end{proof}

We next bound the number of collisions, i.e., $C_j(T)$ in \eqref{eq:upperbound}. 
Under the communication-limited setting, we have the following lemma with respect to the number of collisions $C_j(T)$.

\begin{lemma}\label{lemma:A1-2}
For \cmabucbma, the number of collisions faced by the player in the $N$-best arms in $\cK_b$ by time $T$ is upper bounded as
\begin{align}
      \mathbb{E}[C(T)]\leq N\sum_{k=1}^N\sum_{m=k+1}^K \frac{6\log T}{(\mu_k-\mu_m)^2}+\frac{2\pi^2}{3}KN^2.
\end{align}
\end{lemma}
\begin{proof}
Exactly same procedure as the proof for Lemma \ref{lem:A2_2} is applied. So we omit it.
\end{proof}

Combining the results in Lemma \ref{lemma:A1-1} and Lemma \ref{lemma:A1-2}, and substituting them into \eqref{eq:upperbound}, yields
\begin{align}
R(T)&\leq\mu_1\left(\mathbb{E}[C(T)]+ \sum_{j\in \cK_{-b}}\mathbb{E}[I_{j}(T)]\right)\nonumber\displaybreak[0]\\
 &\leq \mu_1N\Bigg(\sum_{j\in\cK_{-b}}\sum_{k\in\cK\setminus j} \frac{6\log T}{(\mu_k-\mu_j)^2}+\frac{\pi^2}{3}K(K-N)\nonumber\\
 &+\sum_{k=1}^N\sum_{m=k+1}^K \frac{6\log T}{(\mu_k-\mu_m)^2}+\frac{2\pi^2}{3}KN\Bigg)\nonumber\displaybreak[1]\\
    &\leq \mu_1N\Bigg(\sum_{j\in\cK_{-b}}\sum_{k\in\cK\setminus j} \frac{6\log T}{(\mu_k-\mu_j)^2}\nonumber\\
    &+\sum_{k=1}^N\sum_{m=k+1}^K \frac{6\log T}{(\mu_k-\mu_m)^2}+\frac{\pi^2}{3}K(K+N)\Bigg).
\end{align}
This completes the proof of Theorem 2.

\subsection{Auxiliary Lemmas}\label{sec:app-auxiliary}
We provide the following auxiliary lemmas which are used in our proofs.  We omit the proofs of these lemmas for the ease of exposition and refer interested readers to  \cite{Boyd05} , \cite{nedic2009distributed} and \cite{lattimore2020bandit} for details.

\begin{lemma}[Theorem 2 in \cite{Boyd05}]
Assume that $\bP$ is doubly stochastic. 
The limit matrix $\bP^t$ is doubly stochastic and correspond to a uniform steady distribution for all $s$, i.e.,
$$
\lim_{t\rightarrow\infty}\bP^t=\frac{1}{N}\pmb{1}\pmb{1}^T.
$$
\label{themorem_convergence_Phi}
\end{lemma}

\begin{lemma}[Lemma 4 in \cite{nedic2009distributed}]
Assume that $\bP$ is doubly stochastic, the difference between $1/N$ and any element of   $\bP^t$ can be bounded by
\begin{align}
\left|\frac{1}{N}-\bP^t\right|\leq 2 {(1+\beta^{-N})}(1-\beta^{N})^{\frac{t}{N}-1},
\end{align}
where  $\beta$ is the smallest positive value of all consensus matrices, i.e., $\beta=\arg\min P_{i,j}$ with $P_{i,j}>0, \forall i,j.$
\label{lemma_bound_Phi}
\end{lemma}

\begin{lemma}[Corollary 5.5 in \cite{lattimore2020bandit}]
\label{lem:subG_inequality}
Let $X$ be and $\sigma^2$ sub-Gaussian random variable with $\mathbb{E}[X]=\mu.$ Then, for any $a\geq 0$,
\begin{align*}
    \mathbb{P}(|X-\mu|\geq a)\leq e^{-\frac{a^2}{2\sigma^2}}
\end{align*}
\end{lemma}

\begin{lemma}[Lemma 5.4 and Example 5.6 and in \cite{lattimore2020bandit}]
\label{lemma:bound_G}
i) If a random variable $X$ has a finite mean and $a\leq X\leq b$ almost surely, then $X$ is $\frac{1}{4}(b-a)^2$ sub-Gaussian. ii) Let $X_1, X_2,\ldots, X_n$ be $n$ independent random variables such that $X_i$ is $\sigma_i^2$ sub-Gaussian random variable. Then, $X_1+X_2+\ldots+X_n$ is $(\sigma_1^2+\sigma_2^2+\ldots+\sigma_n^2)$ sub-Gaussian.
\end{lemma}

\end{document}